%% file: main.tex
\documentclass{article}

\usepackage{float}
\usepackage{tkz-tab}
\usepackage{xcolor}
\usepackage{bbm}
\usepackage{enumitem}

\usepackage{microtype}
\usepackage{graphicx}
\usepackage{subfigure}
\usepackage{booktabs} 

\usepackage{hyperref}

\usepackage[accepted]{icml2024}

\usepackage{amsmath}
\usepackage{amssymb}
\usepackage{mathtools}
\usepackage{amsthm}

\usepackage{dsfont}
\usepackage{soul}

\usepackage{csquotes}
\usepackage{amsmath,amssymb,amsthm}

\usepackage[capitalize,noabbrev]{cleveref}

\theoremstyle{plain}
\newtheorem{theorem}{Theorem}[section]
\newtheorem{proposition}{Proposition}[section]

\newtheorem{assumption}{Assumption}

\theoremstyle{definition}

\theoremstyle{remark}

\usepackage[textsize=tiny]{todonotes}

\icmltitlerunning{Conformal Predictions under Markovian Data}

\begin{document}

\twocolumn[
\icmltitle{Conformal Predictions under Markovian Data}




\begin{icmlauthorlist}
\icmlauthor{Frédéric Zheng}{yyy}
\icmlauthor{Alexandre Proutiere}{yyy}

\end{icmlauthorlist}

\icmlaffiliation{yyy}{Division of Decision and Control Systems, EECS KTH and Digital Futures, Stockholm, Sweden}

\icmlcorrespondingauthor{Frédéric Zheng}{fzheng@kth.se}

\icmlkeywords{Machine Learning, ICML}

\vskip 0.3in
]



\printAffiliationsAndNotice{}  







\begin{abstract}
We study the split Conformal Prediction method when applied to Markovian data. We quantify the gap in terms of coverage induced by the correlations in the data (compared to exchangeable data). This gap strongly depends on the mixing properties of the underlying Markov chain, and we prove that it typically scales as $\sqrt{t_\mathrm{mix}\ln(n)/n}$ (where $t_\mathrm{mix}$ is the mixing time of the chain). We also derive upper bounds on the impact of the correlations on the size of the prediction set. Finally we present $K$-split CP, a method that consists in thinning the calibration dataset and that adapts to the mixing properties of the chain. Its coverage gap is reduced to $t_\mathrm{mix}/(n\ln(n))$ without really affecting the size of the prediction set. We finally test our algorithms on synthetic and real-world datasets. 
\end{abstract}

\input{1.introduction}
\input{2.relatedwork}
\input{3.preliminaries}
\input{4.marginal_coverage}

\input{6.data_dropping}

\input{10.applications}
\input{11.conclusion}

\section*{Impact Statement} 
This paper presents work whose goal is to advance the field of Machine Learning. There are many potential societal consequences of our work, none which we feel must be specifically highlighted here.

\section*{Acknowledgements}
This research was supported by the Wallenberg AI, Autonomous Systems and Software Program
(WASP) funded by the Knut and Alice Wallenberg Foundation, the Swedish Research Council (VR), and Digital Futures.
\newpage

\bibliography{biblio}
\bibliographystyle{icml2024}

\newpage
\onecolumn
\input{9.proofs}

\input{appendix-sec4}

\end{document}

%% file: 1.introduction.tex
\section{Introduction}

Machine Learning (ML) algorithms and in particular those based on deep neural networks are being increasingly used by practitioners in critical decision processes. If the resulting models often achieve unprecedented levels of prediction performance, they most often come without any guarantees, which can be too hazardous in application fields where safety is crucial (including health care, self driving vehicles, etc.). Introduced and developed by Vovk, see \cite{vovk2005algorithmic,gammerman2007,shafer2008}, Conformal Prediction (CP) is a robust and flexible tool to quantify and handle the inherent uncertainty of these models. The CP framework allows us to build, from any black-box prediction model, a new model whose output is a prediction {\it set} with guaranteed level of certainty. CP has become very popular recently, see  \cite{lei2014distribution,lei2018distribution,romano2019, angelopoulos2023} and references therein, and applied to various learning tasks including regression and classification.

CP typically works as follows. Consider a model $\hat{\mu}$ fitted to a training dataset and with input (or covariate) $X$ and output $\hat{\mu}(X)$. Based on this model and on a {\it calibration} dataset of $n$ (covariate, response) pairs $\{(X_1,Y_t)\}_{t=1}^n$, CP builds, for a given certainty level $1-\alpha$, $C_n(X_{n+1})$ a set of most likely responses. This set is chosen with size as small as possible while ensuring the following coverage guarantee $\mathbb{P}[Y_{n+1}\in C_n(X_{n+1}]\ge 1-\alpha$. 

The coverage guarantees achieved using CP crucially require the calibration data to be
sampled independently and identically (i.i.d.) or at least to be exchangeable. This assumption does not hold in many real-world datasets, where indeed samples may be highly correlated. Such correlations arise naturally in learning tasks pertaining to time series (e.g. predicting the evolution of market behavior) \cite{gibbs2021adaptive,zaffran2022adaptive} or to more general dynamical systems (e.g. in Reinforcement Learning or adaptive control) \cite{foffano2023, dixit2023}. In this paper, we aim at studying CP methods when the correlations in the data are modelled as a Markov chain ($\{(X_1,Y_t)\}_{t\ge 1}$ is a Markov chain). This model is general and includes the classical scenario where the successive covariates form a Markov chain and where the response $Y_t$ remains independent of $\{(X_s,Y_s)\}_{s\neq t}$ \cite{bresler2020,oliveira2022split}. For this model, we address the following questions. {\it (a) How are the coverage and the prediction set size affected by the correlations in the data? 
(b) How can we adapt the CP methods to minimize the impact of these correlations?}

\textbf{Contributions.}\\ 
1. We first provide a theoretical analysis of marginal coverage and of the prediction set size under the classical split CP method in the case of Markovian data. The additional coverage gap due to the correlations depends on the mixing properties of the underlying Markov chain, and we establish that under generic ergodicity assumptions, this gap scales at most as $\sqrt{t_\mathrm{mix}{\ln(n)\over n}}$ where $n$ is the size of the calibration dataset and $t_\mathrm{mix}$ denotes the mixing time of the Markov chain. We further show that typically, the increase in the size of the prediction interval due to the correlations does not exceed $\sqrt{t_\mathrm{mix}/n}$. 

2.  We then investigate the idea of thinning the calibration dataset so as to alleviate the impact of correlations. We refer to as the $K$-split CP, the resulting method where one in $K$ samples of the calibration dataset is kept. We optimize the value of $K$ to achieve an efficient trade-off between coverage and size of the prediction set. The optimal value depends on the mixing time of the Markov chain, and can be estimated in an online manner without really impacting the coverage. $K$-split CP improves the coverage gap that now scales ${t_\mathrm{mix}\over n\ln(n)}$ and has very little impact on the size of the prediction interval.

3. Our theoretical results are confirmed using numerical experiments, both on synthetic and real-world data (e.g., for the prediction on the EUR/SEK exchange rate).

%% file: 2.relatedwork.tex
\section{Related Work}




Extending the CP framework to non-exchangeable data has been considered in the literature, e.g. to model distribution shifts \cite{tibshirani2019, barber2023conformal}. The work \cite{oliveira2022split} is close to ours. There, the authors investigate the Split CP for dependent data verifying a set of concentration-type assumptions, such as stationary $\beta$-mixing chains. They apply concentration inequalities on an independent calibration dataset and come back to the original trajectory by adding an additional $\beta$ coefficient. Our analysis of split CP is inspired by their work. However, we leverage the Markovian nature of the data and in particular concentration results available for Markov chains. The key differences between \cite{oliveira2022split} and our work are: 1) our analysis can be conducted without the stationarity assumption (which we believe is very restrictive); 2) we get more explicit, more generic and simpler upper bounds on the coverage gap for Markovian data; 3) we present an analysis of the size of the prediction interval; 4) we present and investigate the idea of thinning the calibration dataset.


A different line of research consists in adjusting the empirical quantile level to account for the possible undercoverage due to the correlations. For example the authors of \cite{gibbs2021adaptive,zaffran2022adaptive} propose that for each new observed covariate $X_n$, the quantile level $\alpha_n$ used to compute the next CP interval is modified ($\alpha_n$ is updated following a classic stochastic approximation scheme). 


%% file: 3.preliminaries.tex
\section{Preliminaries}

In this section, we first introduce our models, and provide a few existing results on Markov chains and $\beta$-mixing stochastic processes that will be useful in our analysis\footnote{Refer to \cite{meyn2012markov} for a more extensive exposition on Markov chains with general state space, and to \cite{levin2017markov} for mixing properties of Markov chains.}. We then recall the classical Conformal Prediction framework for i.i.d. or exchangeable data. 

\subsection{Markovian data}

We assume that $\{(X_t,Y_t)\}_{t\ge 1-N}$\footnote{Starting the process at time $-N+1$ will turn convenient notation-wise. The first samples up to time 0 will constitute the training data, and the remaining samples the calibration data.} is an homogeneous Markov chain with kernel $P$. The covariates $\{ X_t\}_{t\ge 1-N}$ take values in $\mathcal{X}\subset \mathbb{R}^d$, and the responses $\{ Y_t\}_{t\ge 1-N}$ in ${\cal Y}\subset \mathbb{R}^r$. We assume that the Markov chain is $\phi$-irreducible, aperiodic, and admits a stationary distribution $\pi$. We define $Z_t=(X_t,Y_t)$ and ${\cal Z}={\cal X}\times {\cal Y}$. 

{\bf Example 1.} An important example where $Z=\{Z_t\}_{t\ge 1-N}$ is indeed a Markov chain is the case studied in \cite{bresler2020} where $X=\{X_t\}_{t\ge 1-N}$ is an homogeneous Markov chain of kernel $P_X$, and where given $X_t=x$, the response $Y_t$ is independent of $\{Z_s\}_{s\neq t}$ and of distribution $b(x,\cdot)$. For instance, one may assume that $Y_t=\mu(X_t)+\varepsilon_t$ where $\mu$ is an unknown function and the noise process $\{\varepsilon_t\}_{t\ge 1-N}$ is i.i.d. and independent of $\{X_t\}_{t\ge 1-N}$. The kernel of $Z$ is simply $P_X(x,dx')b(x',dy)$ and its stationary distribution $\pi_X(dx)b(x,dy)$ where $\pi_X$ is that of $X$. Our results apply to this example but to the more general case where $Z$ is a Markov chain.


\textbf{Mixing time.} The mixing time of the Markov chain quantifies the time it takes to approach its steady state. When its transition kernel is $P$, the mixing time is defined as $t_\textrm{mix}=\tau(\frac{1}{4})$ where for any $\varepsilon>0, \tau(\varepsilon)=\min\{t\ge 1:  \sup_z ||P^t(z,.)-\pi||_{TV}\leq \varepsilon\}$. 

\textbf{Geometric ergodicity.} Consider a $\phi$-irreducible and aperiodic Markov chain with  stationary distribution $\pi$. The chain is geometrically ergodic \cite{roberts1997geometric} if there exists a constant $0\le \rho<1$, referred to as the rate of the chain, such that for $\pi$-a.e. $z\in\mathcal{Z}$, there exists $Q(z) <\infty$ with for all $n\ge 1$, $\| P^n(z,.)-\pi\|_{TV}\le Q(z)\rho^n$. We know \cite{nummelin1982geometric} that the function $Q$ can be chosen so that it is $\pi$-integrable ($\int_{\cal Z}Q(z)\pi(dz)<\infty$). The chain is \textit{uniformly} geometrically ergodic if the constant $Q(z)$ is independent of $z$. 

Note that if the state space ${\cal Z}$ is finite, an aperiodic and irreducible Markov chain is uniformly geometrically ergodic \cite{levin2017markov}. Also observe that in the case of Example 1, if $X$ is geometrically ergodic with rate $\rho$, then so is $Z$ with the same rate.

\textbf{Connection between $\rho$ and $t_\mathrm{mix}$.} 
These two quantities, involved in the statement of our results, are closely linked in the case of uniformly geometric Markov chains, as shown in \cite{paulin2015concentration} (see Proposition 3.4). Indeed, the rate $\rho$ of a uniformly geometric ergodic Markov chain can be chosen so that $\rho\le \sqrt{1-\gamma_{\mathrm{ps}}}$ where $\gamma_{\mathrm{ps}}$ denotes its pseudo-spectral gap. Combined with the fact that $\gamma_{\mathrm{ps}}\ge \frac{1}{2t_\mathrm{mix}}$, we obtain $\rho\leq\sqrt{1-\frac{1}{2t_\mathrm{mix}}}$.  When the chain is reversible, it admits an absolute spectral gap $\gamma\geq\gamma_\mathrm{ps}$, leading to a tighter bound of $\rho\leq 1-\gamma$. For a more exhaustive discussion on the connection between the mixing time, the rate and the spectral gap or its variants for non-reversible Markov chain, refer to \cite{chatterjee2023}.

\subsection{$\beta$-mixing processes}\label{subsec:mixing} 

As in \cite{oliveira2022split}, we are interested in $\beta$-mixing processes as they can be divided into approximately independent blocks. This property will be instrumental the analysis of split CP. Let $\{ Z_t\}_{t\ge 1}$ be a stochastic process. For any $s>t\ge 1$, we denote by $Z_s^t=(Z_s,Z_{s+1},\ldots, Z_t)$, by $\mathbb{P}_s^t$ its distribution, and by $\sigma(Z_s^t)$  the $\sigma$-algebra generated by $Z_s^t$. The $\beta$-coefficients of the process are defined as 
$\beta(a)=\sup_{t\geq1}\mathbb{E}\sup \{|\mathbb{P}(B|\sigma(Z_{1}^t))-\mathbb{P}(B)|, B\in\sigma(Z_{t+a}^\infty)\}$ \cite{davydov1973mixing,yu1994rates}. The process is called $\beta$-mixing if $\beta(a)\to 0$ as $a\to\infty$.   
When the process admits a stationary distribution $\pi$, we will also use a slightly different definition of the $\beta$-coefficients: $\beta'(a)=\sup_{t\ge 1} \mathbb{E}[\| \mathbb{P}_{t+a}[\cdot|Z_1^t]-\pi\|_{TV} ]$, where $\mathbb{P}_{t+a}[\cdot|Z_1^t]$ is the conditional distribution of $Z_{t+a}$ given $Z_1^t$. 

The connection between the rate of convergence of Markov chains and their $\beta$-mixing coefficients have been extensively studied. We cite here results from  \cite{davydov1973mixing, nummelin1982geometric, liebscher2005towards} that will turn instrumental in our analysis.





{\it Stationary Geometrically Ergodic Markov chains.} Let $\{ Z_t\}_{t\ge 1}$ be a geometrically ergodic Markov chain with rate $\rho<1$, and with stationary distribution $\pi$. Assume that it is in steady-state ($Z_1\sim\pi$), then its $\beta$-mixing coefficients satisfy \cite{davydov1973mixing}: $\beta(a)\le C\rho^a$, with $C=\int_{\cal Z}Q(z)\pi(dz)$.


{\it Non-stationary Geometrically Ergodic Markov chains.} Let $\{ Z_t\}_{t\ge 1}$ be a geometrically ergodic Markov chain with rate $\rho$, and with stationary distribution $\pi$. Let $\nu_1$ be its initial distribution and suppose  that $\int_\mathcal{Z}Q(z)\nu_1(dz)<\infty$. Then we can show as in the proof of Proposition 3 in \cite{liebscher2005towards} that for all $a\ge 1$, $\beta'(a)\le 3C'(\sqrt{\rho})^{a}$ where $C'=\max(\int_\mathcal{Z}Q(z)\pi(dz), \int_\mathcal{Z}Q(z)\nu_1(dz))$.

\subsection{Split Conformal Prediction} 

We outline below the classical Split CP framework and its basic coverage guarantees in the case of i.i.d. or exchangeable data. Assume that we have access to a dataset $\{(X_t,Y_t)\}_{t=1-N}^{n}$. Given a new data point $X_{n+1}$, we wish to create a conformal prediction set $C_n(X_{n+1})$ with minimal size such that $Y_{n+1}$ lies in $C_n(X_{n+1})$ with probability at least $1-\alpha$.

In the original split CP framework, the dataset is divided into two parts in a random split as the data is exchangeable. However to handle temporal dependencies in our analysis, we use a sequential split: the training dataset is $D_{\textrm{tr}}=\{(X_t,Y_t)\}_{t=1-N}^{0}$ and the calibration dataset $D_{\textrm{cal}}=\{(X_t,Y_t)\}_{t=1}^{n}$. The first dataset is used to fit model $\hat{f}_N: {\cal X}\to \mathbb{R}^r$. Based on this model, a score function $s: {\cal X}\times \mathbb{R}^r\to\mathbb{R}$ is designed (e.g., this function could simply be $s(x,y)= \| \hat{f}_N(x)-y\|$). Next, we compute the scores of the samples in the calibration dataset $s_1=s(X_1,Y_1), \ldots,s_n=s(X_n,Y_n)$, as well $\hat{q}_{\alpha,n}$ as the ${\lceil (n+1)(1-\alpha)\rceil\over n}$ quantile of these scores. Given the new data point $X_{n+1}$, the conformal prediction set $C_n(X_{n+1})$ is finally defined as $\{y : s(X_{n+1},y)\le \hat{q}_{\alpha,n}\}$.

In the case of i.i.d. (or exchangeable) calibration data (i.e., $\{ (X_i,Y_i)\}_{i=1}^{n+1}$ are i.i.d.), the above construction enjoys the following marginal coverage guarantee \cite{vovk2005algorithmic}: 
\begin{equation}\label{eq:gar-iid}
1-\alpha\le \mathbb{P}[Y_{n+1} \in C_n(X_{n+1})] \le 1-\alpha +{1\over 1+n}.
\end{equation}

We investigate whether this guarantee holds in the case of Markovian data and in the case it does not, how we can adapt the above framework. 



    
    


    



\subsection{Notation}

The notation $v_{N,m}={\cal O}_{N,n}(w(N,n))$ means that there exist $M>0$ and two integers $N_0, n_0$ such that for all $N\geq N_0\vee \log^2(n)$, $n\geq n_0$, $|v_{N,m}|\leq M w(N,n)$. The notation ${\cal O}_n, o_n$ are the usual big-O and little-o notation as $n\to\infty$. $Z(N,n)=o_{\mathbb{P}}(1)$ means that $\lim_{n\to\infty} \lim_{N\to\infty}\mathbb{P}[|Z(N,n)| \ge \epsilon] = \lim_{N\to\infty}\lim_{n\to\infty} \mathbb{P}[|Z(N,n)| \ge \epsilon] = 0$ for all $\epsilon>0$. Let ${\cal L}(A)$ be the Lebesgue measure of the set $A$ and $A\Delta B$ be the symmetric difference between sets $A, B$.  

%% file: 4.marginal_coverage.tex
\section{Split CP for Markov Chains}

In this section, we apply the classical Split CP framework to Markovian data, and evaluate the coverage gap induced by the non-exchangeability of the data. We also study the size of the conformal prediction set. 

\subsection{Marginal coverage}

We formally define the notion of {\it coverage gap} as follows. Let $C_n(X_{n+1})$ be a prediction set based on the $n$ last observed samples taken from the calibration dataset. We say that it has a coverage gap smaller than $\gamma\ge 0$ if:
\begin{equation}\label{eq:gap}
\mathbb{P}[Y_{n+1} \in C_n(X_{n+1})] \in \left[1-\alpha-\gamma, 1-\alpha+\gamma+{1\over n}\right].
\end{equation}

Consider the Markov chain $\{(X_t,Y_t)\}_{t\ge -N+1}$ with initial distribution $\nu_0$ (i.e., $(X_{1-N},Y_{1-N})\sim \nu_0$) and stationary distribution $\pi$. In the following, to simplify the notation, we define $\delta(a)=\|\nu_0P^a-\pi\|_{TV}$ for any $a\in \mathbb{N}$. When applying the standard Split CP to this Markov chain, the induced coverage gap will necessarily depends on its transient behavior, partly described by $t_\mathrm{mix}$ and by the sequences $\{\delta(a)\}_{a\ge 0}$ and $\{\beta(a)\}_{a\ge 0}$. To analyze the coverage gap, we consider two scenarios:

{\it 1. With restart.} In this scenario, we assume that the calibration and the training datasets are independent, and that the Markov chain is restarted with distribution $\nu_1$ at time 1. Assuming such independence makes the conditioning on the training dataset easy and simplifies the analysis. When applying a restart, we will use the notation: for $a\ge 0$, $\delta_1(a)=\|\nu_1P^a-\pi\|_{TV}$. 

{\it 2. Without restart.} Here, the samples composing the training and calibration datasets come from a single trajectory of the Markov chain. The datasets are hence not independent and the distribution of $(X_1,Y_1)$ is $\nu_0P^N$. We tackle this case by creating a separation of $r\in[n]$ time-steps between the training and calibration datasets so as to reduce their stochastic dependence.

The following proposition provides upper bounds on the coverage gap in both scenarios. These upper bounds will be optimized in various cases depending on the nature of the Markov chain. 

\begin{proposition}\label{prop:main} 
(1. With restart) Applying Split CP with restart yields a coverage gap $\gamma$ satisfying, for any $u>{1\over n}\sum_{a=1}^n \delta_1(a)$, $\gamma\le \gamma(u)$ where
\begin{equation}\label{eq:restart}
\gamma(u)= u 
 + e^{-{2n\over 9t_\mathrm{mix}}\big(u-{1\over n}\sum_{a=1}^n \delta_1(a)\big)^2} + \delta_1(n+1).   
\end{equation}
(2. Without restart) Applying Split CP yields a coverage gap $\gamma$ satisfying, for any $u> \delta(N)$ and for any $r\in [n]$, $\gamma\le \gamma(u,r)$ where
\begin{align}
    \gamma(u,r) = u + & e^{-{2(n-r)\over 9t_\mathrm{mix}}\big(u-\delta(N)\big)^2} + \delta(n+N+1)\nonumber\\
    &+2\beta(r)+{1+\alpha r\over n+1}.\label{eq:gamma_gen}
\end{align}
\end{proposition}

We may optimize the above coverage gap upper bounds over $u$ and $r$. The following theorem summarizes the outcomes of this optimization.

\begin{theorem}\label{thm:main}
(1. With restart) Applying Split CP with restart yields a coverage gap $\gamma$ satisfying
\begin{equation}\label{eq:restart2}
\gamma = {\cal O}_n\left(\sqrt{t_\mathrm{mix}{\ln(n)\over n}} + \delta_1(n+1) + {1\over n}\sum_{a=1}^n \delta_1(a)\right).   
\end{equation}
We deduce that, if the Markov chain is geometrically ergodic, then $\gamma = {\cal O}_n(\sqrt{t_\mathrm{mix}{\ln(n)\over n}})$.\\
(2. Without restart) Applying Split CP to a geometrically ergodic Markov chain yields a coverage gap $\gamma$ satisfying
\begin{equation}\label{eq:gamma_gen2}
\gamma = {\cal O}_{N,n}\left(\sqrt{t_\mathrm{mix}{\ln(n)\over n}}\right).
\end{equation}
\end{theorem}

Proposition \ref{prop:main} and Theorem \ref{thm:main} quantify the coverage gap due to the non-exchangeability of the data. The gap depends (in both cases with or without restart) on the mixing properties of the underlying Markov chain. As expected, the gap becomes negligible only if the number of samples in the calibration dataset significantly exceeds the mixing time. 









\subsection{Size of the conformal prediction set}\label{subsec:size}

To investigate the impact of the non-exchangeability of the data on the size of the conformal prediction set, we use the same setting as that used in \cite{lei2018distribution}. We consider real valued responses, and we make the following assumptions (similar to those made in Example 1).

\begin{assumption}
\label{assumption:score}
For any $t\ge -N+1$ and $x$, given $X_t=x$, the response $Y_t$ can be written as $\mu(x)+\varepsilon$ where the density of the noise $\varepsilon$ is symmetric around 0 and non-increasing on $\mathbb{R}_+$. The training dataset is used to design an estimator $\hat{\mu}_N$ of the function $\mu$, and the score function used in the CP method is defined through the residuals: $s_t=|Y_t -\hat{\mu}_N(X_t)|$.
\end{assumption}

The next assumption states that the estimator $\hat{\mu}_N$ becomes accurate as the size of the training dataset increases. This assumption is not critical, and results for the size of the prediction set can also be obtained by just assuming {\it stability} \cite{lei2018distribution}, i.e., convergence of $\hat{\mu}_N$ to a given $\tilde{\mu}$ possibly different than $\mu$.  

\begin{assumption}
\label{assumption:stability} There exist two sequences $\{c_N\}_{N\ge 1}$ and $\{d_N\}_{N\ge 1}$ such that $c_N=o_N(1), d_N=o_N(1)$ and\\ $\mathbb{P}(||\hat{\mu}_N-{\mu}||_\infty \geq c_N) \leq d_N.$
\end{assumption} 

\begin{assumption}
\label{assumption:density}
   The noise $|Y-{\mu}(X)|$ admits a density ${f}$ that is lower bounded by $\kappa>0$ on a neighborhood ${\cal N}$ of its $(1-\alpha)$ quantile ${q}_\alpha$. 
\end{assumption}

Under the above assumptions and should the data be i.i.d., we know \cite{lei2014distribution} that the optimal conformal prediction set (that with conditional coverage $1-\alpha$ and minimal size) is given $X_{n+1}=x$, $C^\star(x)=[\mu(x)-q_\alpha,\mu(x)+q_\alpha]$, where $q_\alpha$ is the $(1-\alpha)$ quantile of the law of $|\varepsilon|$. We use the set $C^\star(x)$ as a benchmark to study the impact of (i) non-i.i.d. data and (ii) of the fact that $\mu$ and $q_\alpha$ are unknown. We compare $C^\star(x)$ to the returned conformal prediction set $C_n(x)=[\hat{\mu}_N(x)-\hat{q}_{\alpha,n}, \hat{\mu}_N(x)-\hat{q}_{\alpha,n}]$.

\subsubsection{Concentration of empirical quantiles}

We start quantifying the deviation between the empirical quantile $\hat{q}_{\alpha,n}$ used to build the prediction set and its true counterpart $\hat{q}_{\alpha}$, the $(1-\alpha)$ quantile of $|Y-\hat{\mu}_N(X)|$ when $(X,Y)\sim \pi$. We introduce the set ${\cal U}=\{t>0 : [{q}_{\alpha-t}, {q}_{\alpha+t}]\subset {\cal N} \}$. 

\begin{proposition}
\label{prop:length}
Suppose that Assumptions \ref{assumption:score}, \ref{assumption:stability}, \ref{assumption:density} hold.\\
(1. With restart) Let $u\ge 0$ and let $u'=u-(2\kappa c_N+d_N)$. Assume that $u'$ lie in ${\cal U}$, and that $u'>{1\over n}\sum_{a=1}^n\delta_1(a)$. Then we have:
\begin{equation*}
    \label{eq:length_rR}
         \mathbb{P}(|\hat{q}_{\alpha,n}-\hat{q}_{\alpha}|> \frac{u}{\kappa})
     \le 2e^{-\frac{2n}{9t_\mathrm{mix}}\big(u'-{1\over n}\sum_{a=1}^n\delta_1(a)\big)^2} + d_N.
    \end{equation*}
Therefore, for any $\delta>0$ with probability at least $1-\delta-d_N$, we have $|\hat{q}_{\alpha,n}-\hat{q}_{\alpha}| \le {u^\star\over \kappa}$ with
$$
u^\star=d_N + 2\kappa c_N+ \frac{1}{n}\sum_{i=1}^n\delta_1(a)+\sqrt{\frac{9t_\mathrm{mix}\ln (\frac{2}{\delta})}{2n}},
$$
provided that $u^\star\in {\cal U}$. \\
(2. Without restart) For any $r\in[n]$ and $u\in\mathbb{R}$, let $u'=u+\frac{1+r\alpha}{n+1}-(2\kappa c_N+d_N)$. Assume that both $\frac{1+r\alpha}{n+1}$ and $u'$ lie in ${\cal U}$ and that $u'>\delta(N)$. Then we have:
    \begin{multline*}
    \label{eq:length_r}
         \mathbb{P}(|\hat{q}_{\alpha,n}-\hat{q}_{\alpha}|> \frac{u}{\kappa})
     \le 2e^{-\frac{2(n-r)}{9t_\mathrm{mix}}\big(u'-\delta(N)\big)^2} +d_N+\beta(r).
    \end{multline*}
Further assuming that the Markov chain is geometrically ergodic with rate $\rho$, we deduce that for any $\delta>0$ with probability at least $1-\delta-d_N-\frac{2}{n}$, we have $|\hat{q}_{\alpha,n}-\hat{q}_{\alpha}| \le {u^\star\over \kappa}$ with 
$$
u^\star=d_N + 2\kappa c_N+ \delta(N)+\alpha\frac{\ln (n)}{n\ln{(\frac{1}{\rho})}}+\sqrt{\frac{9t_\mathrm{mix}\ln (\frac{2}{\delta})}{2n}},
$$
provided that $u^\star\in {\cal U}$. 



    \end{proposition}

Note that when the Markov chain is geometrically ergodic, with or without restart, we indeed obtain that $\hat{q}_{\alpha,n}$ concentrates around $\hat{q}_{\alpha}$ as $n$ and $N$ grow large. Further observe that as a direct application of the above proposition, we recover the result from \cite{kolla2019concentration} stating that in the case of i.i.d. data, with probability at least $1-\delta$, $|\hat{q}_{\alpha,n}-\hat{q}_{\alpha}| \le c\sqrt{\ln ({2}/{\delta})/2n}$ where $c$ is a constant that depends on $\kappa$. Handling Markovian data induces an additional multiplicative factor $\sqrt{t_\mathrm{mix}}$ in the difference between the true and the estimated quantile of the residuals. 


\subsubsection{Asymptotic optimality}

To compare $C_n(x)$ to $C^\star(x)$, we further need to make an assumption about the consistency of the estimated model $\hat{\mu}_N$. We make the same assumption as in \cite{lei2018distribution}: 

\begin{assumption} 
\label{assumption:consistence}The estimated model $\hat{\mu}_N$ satisfies: $\Delta_N=\mathbb{E}_{X\sim \pi, \mathrm{tr}}[(\hat{\mu}_N(X)-\mu(X))^2]=o_N(1)$.    
\end{assumption}

As shown in \cite{lei2018distribution}, this assuption allows us to control the gap between the true quantile $q_\alpha$ and that of the estimated model $\hat{q}_\alpha$. Indeed, if we assume that the density function of $|\varepsilon |$ has continuous derivate upper bounded by $M>0$ and is lower bounded by $\kappa>0$ on $(q_\alpha -\eta, q_\alpha+\eta)$ for some $\eta > (M/2\kappa)\Delta_N$, then $|\hat{q}_\alpha - q_\alpha|\le (M/2\kappa)\Delta_N$. Combining this result to that of Proposition \ref{prop:length}, we can finally quantify the difference between $C_n(X_{n+1})$ and $C^\star(X_{n+1})$.

\begin{proposition}\label{opt:length} Assume that the Markov chain $Z=(X,Y)$ is $\beta$-mixing. We have:
$$
{\cal L}(C^\star(X_{n+1})\Delta C_n(X_{n+1}))=o_{\mathbb{P}}(1),
$$
if either (i) the Markov chain is in steady-state (i.e., $\nu_0=\pi$ without restart or $\nu_1=\pi$ with restart), and Assumptions \ref{assumption:score}, \ref{assumption:density}, \ref{assumption:consistence} hold, or (ii) Assumptions \ref{assumption:score}, \ref{assumption:stability}, \ref{assumption:density}, \ref{assumption:consistence} hold.
\end{proposition}

The proposition states that asymptotically, when the calibration dataset grows large, the prediction set becomes very close to the optimal conformal prediction set. It is worth noting that we do not assume that the Markov chain is geometrically ergodic. We do not quantify the speed at which $C_n(X_{n+1})$ converges to $C^\star(X_{n+1})$, and hence assuming $\beta$-mixing is enough.

%% file: 6.data_dropping.tex
\section{$K$-Split Conformal Prediction}

The coverage gap $\gamma$ identified when applying the classical CP approach to Markovian data stems from the correlations in the  data. A natural way to decrease the impact of these correlations consists in thinning the calibration data. Specifically, we may build the prediction set based on calibration samples taken every $K$ steps. We refer to this approach as the {\it $K$-split conformal prediction}. It applies the classical CP method to the calibration data $D_{\textrm{cal},K}=\{(X_{iK+1},Y_{iK+1})\}_{i=0}^{\lfloor \frac{n}{K}\rfloor-1}$. In the following, for simplicity, we assume that $n=Km$ for some $m\in \mathbb{N}$ so that $D_{\textrm{cal},K}$ consists of $m$ samples. We define $\hat{q}_{\alpha,K,n}$ as the ${\lceil (m+1)(1-\alpha)\rceil\over m}$ quantile of the scores $s_1=s(X_1,Y_1), \ldots,s_m=s(X_{K(m-1)+1},Y_{K(m-1)+1})$. Given the new data point $X_{n+1}$, the conformal prediction set $C_{K}(X_{n+1})$ is defined as $\{y : s(X_{n+1},y)\le \hat{q}_{\alpha,K,n}\}$.

The analysis made in the previous section extends here since the thinned process remains a Markov chain but with transition kernel $P^K$. However, applying Theorem \ref{thm:main} does not provide interesting results. Indeed, observe that the mixing time of thinned process is $t_{\mathrm{mix},K}=\lceil {t_{\mathrm{mix}}\over K}\rceil$. This implies that the coverage gap $\gamma$ obtained in Theorem \ref{thm:main} would scale as 
$$\sqrt{\lceil {t_{\mathrm{mix}}\over K}\rceil K{\ln(n/K)\over n}}\ge \sqrt{t_{\mathrm{mix}}{\ln(n/K)\over n}}.
$$
Hence, from this analysis, we do not see any improvement in thinning the process. We propose below an alternative analysis based on the so-called {\it blocking technique} (that can be traced back to \cite{Bernstein1927SurLD}).

\subsection{Marginal coverage guarantees via the blocking technique}

The blocking technique applies not only to Markov chains but also to $\beta$-mixing processes. Hence we start by stating general results for these processes, and then specialize the results in the case of Markov chains.

\subsubsection{$\beta$-mixing processes}

Let $\{X_t,Y_t\}_{1-N}^n$ a stochastic process with initial distribution $\nu_0$ and stationary distribution $\pi$. For any $r\in[n]$, denote by $\beta'_{r}(a)$ the mixing coefficient calculated on the trajectory $\{(X_t,Y_t)\}_{t=r}^{n}$, i.e., $\beta'_{r}(a)=\sup_{t\ge r} \mathbb{E}[\| \mathbb{P}^{t+a}[\cdot|(X_i,Y_i)_{i=r}^{t}]-\pi\|_{TV} ]$.

\begin{proposition}
    \label{proposition:cov_k_split}
    (1. With restart) Under the $K$-Split CP with restart, $\mathbb{P}[Y_{n+1}\in C_{K}(X_{n+1})]$ belongs to the interval $[1-\alpha-\gamma(K), 1-\alpha+\gamma(K)+\frac{K}{n}]$ for any $n\ge 1$ and any $K\in [n]$ where
\begin{align}
\gamma(K)=
\begin{cases}
     2\frac{n}{K}\beta(K) &\text{if $\nu_1=\pi$}, \\
     2\frac{n}{K}\beta'_{1}(K) & \text{otherwise}.
\end{cases}
\end{align}

(2. Without restart) Under the $K$-Split CP without restart, $\mathbb{P}[Y_{n+1}\in C_{K}(X_{n+1})]$ belongs to the interval \\
$[1-\alpha-\gamma(K,r), 1-\alpha+\gamma(K,r)+\frac{K}{n-r}]$ for any $n\ge 1, r\in[n], K\in[n-r]$ where
\begin{align}
\gamma(K,r)=
\begin{cases}
    \frac{1+\alpha r}{n+1}+ 2\frac{n-r}{K}\beta(K)+\beta(r) &\text{if $\nu_0=\pi$}, \\
     \frac{1+\alpha r}{n+1}+ 2\frac{n-r}{K}\beta'_{r}(K)+\beta(r) & \text{otherwise}.
\end{cases}
\end{align}

\end{proposition}

The above bounds on the coverage gap are rather simpler than those obtained in the case without thinning (see Proposition \ref{prop:main}) and only depend on the $\beta$ coefficients of the process. Note that these bounds do make sense only when these coefficients decrease rapidly with $K$. They are also useless in the case without thinning ($K=1$).

The term $(1-\alpha)$ appearing in the lower bounds of the coverage stems from the analysis of split CP in the i.i.d setting. In this setting, due to the finiteness of the calibration dataset, the coverage is actually lower bounded by : 
\begin{equation}
    \mathrm{coverage}\geq  \frac{\lceil (n_\mathrm{cal}+1)(1-\alpha)\rceil}{n_\mathrm{cal}+1}\geq 1-\alpha,
    \label{eq:iid_bound}
\end{equation}
where $n_\mathrm{cal}$ is the size of the calibration dataset which is $\frac{n}{K}$ for $K$-split CP. When $n_\mathrm{cal}$ (resp. $\frac{n}{K}$) is small, the difference between the two bounds can be non negligible and lead to an over-coverage of split CP (resp. $K$-split CP). We illustrate this phenomenon and propose a slight refinement of the analysis of $K$-split CP in Section 6.

\subsubsection{Geometrically ergodic Markov chains}

Next we optimize the value of $K$ to achieve a good trade-off between coverage and size of the prediction set. Observe that indeed such a trade-off exists. If we just want to maximize coverage, then we can select values of $K$ maximizing the functions $\gamma(K)$ or $\gamma(K,r)$ defined in Proposition \ref{proposition:cov_k_split}. This would lead to choosing $K$ very large, but would be at the expense of enlarging the prediction set. This trade-off is confirmed by the terms $K/n$ and $K/(n-r)$ in the upper limit of the intervals where $\mathbb{P}[Y_{n+1}\in C_{K}(X_{n+1})]$ lies. To address the coverage-size trade-off, we choose to minimize the size of these intervals. Specifically, for the case with restart, we wish to minimize $\gamma(K)+K/n$ over $K$; whereas for the case without restart, the function to minimize is $\gamma(K,r)+K/(n-r)$ over $K$ and $r$. 

To this aim, we assume that the process is a geometrically ergodic Markov chain with rate $\rho$. In the case without restart, to simplify the optimization in $r$, we observe that $\beta'_{r}(a) \leq  \beta'_{1}(a) \leq C'\rho^\frac{a}{2}$ for $C'=3\max(\int_\mathcal{X}Q(x)\pi(dx), \int_\mathcal{X}Q(x)\nu_1(dx))<\infty$ (refer to subsection \ref{subsec:mixing}).

\begin{theorem}
    \label{thm:cov_k}
    (1. With restart) Assume $\int_\mathcal{X}Q(x)\nu_1(dx)<\infty$. Under the $K$-Split CP with restart   
    and with\footnote{$W_0$ is the Lambert function of order 0.}\\ $K^\star=\frac{W_0(n^2(\ln \rho)^2)}{\ln(\frac{1}{\rho})}$, 
    $\mathbb{P}(Y_{n+1}\in C_{K^\star}(X_{n+1}))$ belongs to\\ $[1-\alpha-\gamma_\mathrm{low},1-\alpha+\gamma_\mathrm{up}]$ with $\gamma_\mathrm{low}={\cal O}_n(\frac{1}{n\ln (n)\ln(\frac{1}{\rho})})$ and $\gamma_\mathrm{up}={\cal O}_n(\frac{\ln (n)}{n\ln(\frac{1}{\rho})})$.

    (2. Without restart) The same result as in the case of restart holds if we assume $\int_\mathcal{X}Q(x)\nu_0P^N(dx)<\infty$. 
\end{theorem}

This theorem states that for geometrically ergodic Markov chains, the coverage gap of the optimal $K$-split CP scales as $\frac{1}{n\ln (n)\ln(1/\rho)}$. For clarity, we restate the result with $t_\mathrm{mix}$ only as in Section 4. As discussed in Section 3, we can choose $\rho$ satisfying $\rho\le (1-{1\over 2t_\mathrm{mix}})^{1/2}$. Therefore, the coverage gap of optimal $K$-split CP scales as $\frac{t_\mathrm{mix}}{n\ln (n)}$ Recall that the gap of the split CP without thinning was scaling as $\sqrt{t_\mathrm{mix}{\ln(n)\over n}}$, therefore $K$-split CP divides the coverage gap by a factor $\sqrt{n}\ln(n)^{3/2}$.

\subsection{Size of the conformal prediction set}

As mentioned earlier, by thinning the initial Markov chain, we obtain another Markov chain with kernel $P^K$. Hence our results pertaining to the size of the prediction set and derived in Subsection \ref{subsec:size}  remain valid. More precisely, all statements made in Proposition \ref{prop:length} hold provided that we replace $n$ by $n/K$, $t_\mathrm{mix}$ by $\lceil {t_{\mathrm{mix}}\over K}\rceil$, and ${1\over n}\sum_{a=1}^n\delta_1(a)$ by  ${K\over n}\sum_{a=1}^{\lfloor n/K\rfloor}\delta_1(Ka)$ (note that all terms related to the training data, e.g. $c_N,d_N$, remain unchanged). When applying the results to geometrically ergodic chains, we also have to replace the rate $\rho$ by $\rho^K$. The following proposition summarizes the above observations.

\begin{proposition}\label{prop:sizeK}
Suppose that Assumptions \ref{assumption:score}, \ref{assumption:stability}, \ref{assumption:density} hold. Assume that the Markov chain is geometrically ergodic with rate $\rho$. Applying the $K$-Split CP with or without restart yields that: for any $\delta>0$ with probability at least $1-\delta-d_N-\frac{2}{n}$, we have $|\hat{q}_{\alpha,n}-\hat{q}_{\alpha}| \le {u^\star\over \kappa}$ with 
$$
u^\star=d_N + 2\kappa c_N+ \delta(N)+{\cal O}_n\left(\sqrt{\frac{(t_\mathrm{mix}\vee K)\ln (\frac{2}{\delta})}{n}}\right),
$$
provided that $u^\star\in {\cal U}$. 
\end{proposition}

We can plug $K^\star$, the optimal value of $K$ identified in Theorem \ref{thm:cov_k}, in the above result. Using the fact that $W_0(x)=O(\ln x)$, we have $K^\star={\cal O}_n(\frac{\ln(n)}{\ln(\frac{1}{\rho})})={\cal O}_n(t_\mathrm{mix}\ln(n))$. Hence with this choice of $K$, the high probability upper bound of $|\hat{q}_{\alpha,n}-\hat{q}_{\alpha}|$ goes 
\begin{tabular}{rcl}
from & ${\cal O}_n(\sqrt{\frac{t_\mathrm{mix}\ln (\frac{2}{\delta})}{n}})$ & for the classical Split CP\\
to & ${\cal O}_n(\sqrt{\frac{t_\mathrm{mix}\ln(n)\ln (\frac{2}{\delta})}{n}})$ & for the $K^\star$-Split CP.   
\end{tabular}

This result suggests that the impact of thinning the calibration dataset using the optimal $K^\star$ does not impact much the size of conformal prediction set. Finally observe that with thinning, the results of Proposition \ref{opt:length} hold.

\subsection{Adaptive $K$-Split CP}

As demonstrated in Theorem \ref{thm:cov_k}, the choice of the thinning parameter $K$ in the $K$-split CP is critical to achieve an efficient trade-off between coverage and prediction set size. However, $K^\star$ is a function of the rate $\rho$ of the Markov chain, and the latter is initially unknown. We assume below that the state space is finite so that the Markov chain is uniformly geometric ergodic. In this case, its rate can be selected as $\rho=\sqrt{1-\gamma_\mathrm{ps}}$ or $\rho=1-\gamma$ if the chain is reversible. Both quantities $\gamma_\mathrm{ps}$, $\gamma$ can be estimated \cite{hsu2015mixing,hsu2019,combes2019,wolfer2019estimating}. For simplicity, we state the results in the reversible case and assuming that we estimate $\gamma$ using the training dataset. Our results can however be easily extended to the non-reversible case by estimating $\gamma_\mathrm{ps}$ \cite{wolfer2019estimating}.

We use the estimator $\hat{\gamma}_N$ proposed in \cite{hsu2015mixing} to construct $\hat{\rho}_N=1-\hat{\gamma}_N$. This estimator enjoys the following guarantees: for any $N\ge 1$, for any $\delta\in (0,1)$,
$$
\mathbb{P}_\mathrm{tr}\left[|\hat{\rho}_N-\rho|\ge C\sqrt{\frac{\ln (\frac{|\mathcal{X}|}{\delta})\ln (N)}{\pi_\star(1-\lambda) N}}\right]\le \delta,
$$
where $C$ is a universal constant and $\pi_\star:=\min_{z\in {\cal Z}}\pi(z)$ (here $\pi$ is the stationary distribution of the Markov chain). Based on $\hat{\rho}_N$, we build the following estimator for $K^\star$: $\hat{K}_N = \ln(n)/\ln(1/\hat{\rho}_N)$. When plugging this value in our $K$-Split CP method, we get the following guarantees. 

\begin{proposition}
    \label{thm:cov_k2}
    (1. With restart) Assume that $\int_\mathcal{X}Q(x)\nu_1(dx)<\infty$. Under the $K$-Split CP with restart   
    and thinning parameter $\hat{K}_N$,
    $\mathbb{P}(Y_{n+1}\in C_{\hat{K}_N}(X_{n+1}))$\\ belongs to the interval \\
    $[1-\alpha-\gamma_\mathrm{low}-l_{N,n},1-\alpha+\gamma_\mathrm{up}+u_{N,n}]$ with $\gamma_\mathrm{low}, \gamma_\mathrm{up}$ as in Theorem \ref{thm:cov_k} and 
    \begin{align*}
    l_{N,n} & ={\cal O}_{N,n}({\ln(N)\over \sqrt{N}}{1\over \rho(\ln\rho)^2n\ln(n)}),\\
    u_{N,n} & ={\cal O}_{N,n}({\ln(N)\over \sqrt{N}}\frac{1}{n})
    \end{align*}
    (2. Without restart) The same result as in the case of restart holds if we assume $\int_\mathcal{X}Q(x)\nu_0P^N(dx)<\infty$. 
\end{proposition}

From the above results, we conclude that the estimation of $K^\star$ does not impact much the coverage gap, and one can easily enjoy the benefits of the optimal $K$-split CP without the apriori knowledge of $K^\star$.

%% file: 10.applications.tex
\section{Experiments}

We illustrate experimentally the coverage gap and set length of our two described conformal prediction methods, namely original split conformal and $K$-split conformal. The score considered is the usual residual $|Y-\hat{\mu}(X)|$ (all responses are real-valued) therefore the conformal set length is simply $2\hat{q}_{\alpha,n}$ (resp $2\hat{q}_{\alpha,K,n}$). When the model is specified, the latter is compared with the optimal length $2q_\alpha$ (as defined in Section 4.2) through a relative difference. 

The objective of our experiments is to mainly assess the performance of the original split CP method in the case of Markovian data, and that of our proposed algorithms, namely $K$-split CP and corrected $K$-split CP (as defined below). Note that we do not compare our algorithms to those designed to cope with non-stationary dependent data such as ACI \cite{gibbs2021adaptive, zaffran2022adaptive} (ACI extends the original CP algorithm by allowing the confidence level $\alpha$ to change in an active and online manner depending on the observed coverage). 

\subsection{K-split conformal prediction in practice}

As discussed in Section 5, $K$-split CP may exhibit an over-coverage when there are not enough values in the reduced calibration dataset. To circumvent this issue, we can adjust the quantile level used by $K$-split CP. This level is set to $\alpha'$ by taking into account the difference between the two bounds presented (\ref{eq:iid_bound}). Specifically, we select $\alpha'$ such that 

$$
\frac{\lceil (\frac{n}{K}+1)(1-\alpha')\rceil}{\frac{n}{K}+1} = 1-\alpha
$$




\subsection{Synthetic data}

In this subsection, we apply all split CP methods to synthetic examples, one with a finite state space and one with a continuous state space. An experiment consists in generating one trajectory of length $N+n+1$ and in applying CP to the last point. We repeat the experiment $N_\mathrm{trials}=1000$ times and report the average coverage rate. We fix $N=10000$.

\subsubsection{The lazy random walk on $\mathbb{Z}/w\mathbb{Z}$}

Consider the stationary lazy random walk. It is a simple example of a finite state space, irreducible, aperiodic and reversible Markov Chain, for which 
$\rho=\lambda_2=\frac{1+cos(\frac{2\pi} {w})}{2}$. For $w=20$, this example already exhibits strong temporal correlations as $\rho=0.97$. The discrete state space setting allows to use the estimator from \cite{hsu2015mixing}. For a given true model $\mu(x)$ and independent symmetric noise $\varepsilon_t$, we generate $Y_t=\mu(X_t)+\varepsilon_t$ (refer to Appendix \ref{app:lazy_walk} for details).

\begin{figure}[ht]
\vskip 0.2in
\begin{center}
\centerline{\includegraphics[width=\columnwidth]{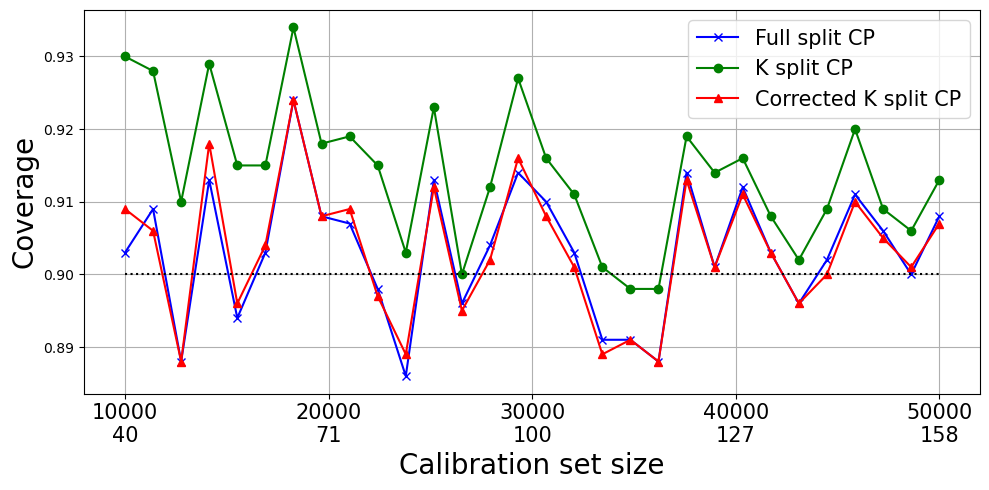}}
\caption{Coverage for the lazy walk ($w=20$) as a function of $n$. The second values on the $x$-axis represent the (optimal) values of $K^\star(n)$.  
}
   \label{fig:cov_lazy}
\end{center}
\vskip -0.2in
\end{figure}
\begin{figure}
\begin{center}
\centerline{\includegraphics[width=\columnwidth]{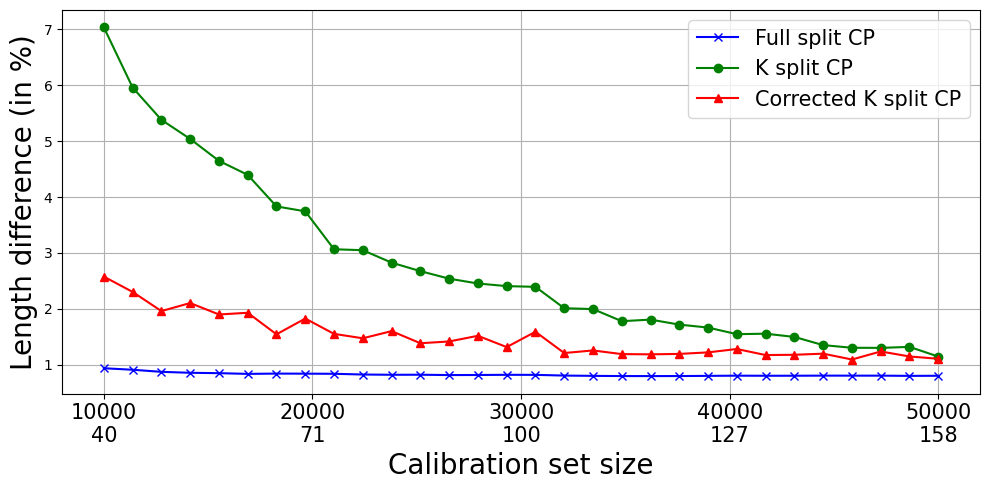}}
    \caption{$\frac{|q_{n,\alpha}-q_\alpha|}{q_\alpha}$ for the lazy walk as a function of $n$. The second values on the $x$-axis represent the (optimal) values of $K^\star(n)$. }
    \label{fig:length_lazy}
    \end{center}
\vskip -0.2in
\end{figure}

In Figure \ref{fig:cov_lazy}, we observe that all three methods achieve almost $1-\alpha$ coverage gap as $n$ increases. Note that $K$-split CP always has stronger coverage than full split CP, and overcovering steadily diminishes when the number of samples in the reduced dataset increase. The correction proposed allows to remain closer to $1-\alpha$ in the regime where $n$ is small as expected. Figure \ref{fig:length_lazy} shows that all CP methods output an interval whose length approaches the optimal length as $n$ increases. Again, we observe the overcoverage phenomenon when $n$ is low for $K$-split CP, but asymptotically vanishes as analysed in Section 5.

\subsubsection{The Gaussian autoregressive (AR) model of order 1}

The classical AR(1) models are reversible Markov chains defined by the following recursive equation: $\forall n$, $X_{n+1}=\theta X_n + \varepsilon_{n+1}$ with $\varepsilon_n \sim N(0,\omega^2)$ and for some $\theta\in[0,1[$ and $\omega>0$. $\varepsilon_n$ is a Gaussian noise independent of $X$. For this stochastic process, we predict the AR model itself, i.e., $Y_t=X_{t+1}$. $(X,Y)$ is a Markov chain with the same kernel as $X$ and it is geometrically ergodic \cite{bhattacharya1995ergodicity}. Since the state space is continuous, we can not use the estimator defined in Section 5.3, however in this example, we can explicity compute $\rho$ (refer to Appendix \ref{app:gaussian_ar} for details).

\begin{figure}[ht]
\vskip 0.2in
\begin{center}
\centerline{\includegraphics[width=\columnwidth]{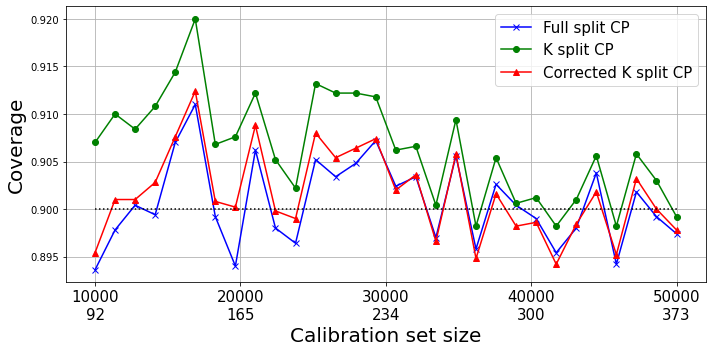}}
    \caption{Coverage for Gaussian AR $(\theta=0.9, \omega=1)$ as a function of $n$. The second values on the $x$-axis represent the (optimal) values of $K^\star(n)$. }
    \label{fig:cov_ar}
    \end{center}
\vskip -0.2in
\end{figure}

In Figure \ref{fig:cov_ar}, we compare our different CP methods. Again, we observe that $K$-split CP outperforms full split CP. But all methods achieve a very good coverage, close to $1-\alpha$ as $n$ grows large.



\subsection{Real-world application}

In this subsection, we apply split CP to real-world datasets. Experiments to assess the performance of split CP on dependent data have already been conducted in \cite{wisniewski2020application} (without guarantees) and in \cite{oliveira2022split} (with guarantees for stationary $\beta-$mixing data). We further consider non-stationary data and investigate the performance of $K$-split CP.

\subsubsection{Exchange rate EUR/SEK}

The objective here is to predict the exchange rate EUR/SEK\footnote{The dataset can be found at \url{https://www.histdata.com/}} in 2022. Let $X_t$ be the exchange rate, reported every minute. As done in \cite{oliveira2022split}, we assume that the series of returns $r_t=\frac{X_{t+1}}{X_t}-1$ is $\beta$-mixing with exponentially fast convergence to 0. We estimate the corresponding value $\rho$ by computing the auto-correlations as they exhibit a similar decay as that of the $\beta$ coefficients, and applying a simple linear regression to their logarithms. This gives a value of $\rho=0.57$.

At each timestep, we apply conformal prediction to the next return $r_{t+1}$ with a rolling window of fixed size divided into training and calibration datasets (1 month = 30x24x60 data points for each in this example). We compute whether the given CP method covered $r_{t+1}$ in 2022.

\begin{figure}[ht]
\vskip 0.2in
\begin{center}
\centerline{\includegraphics[width=\columnwidth]{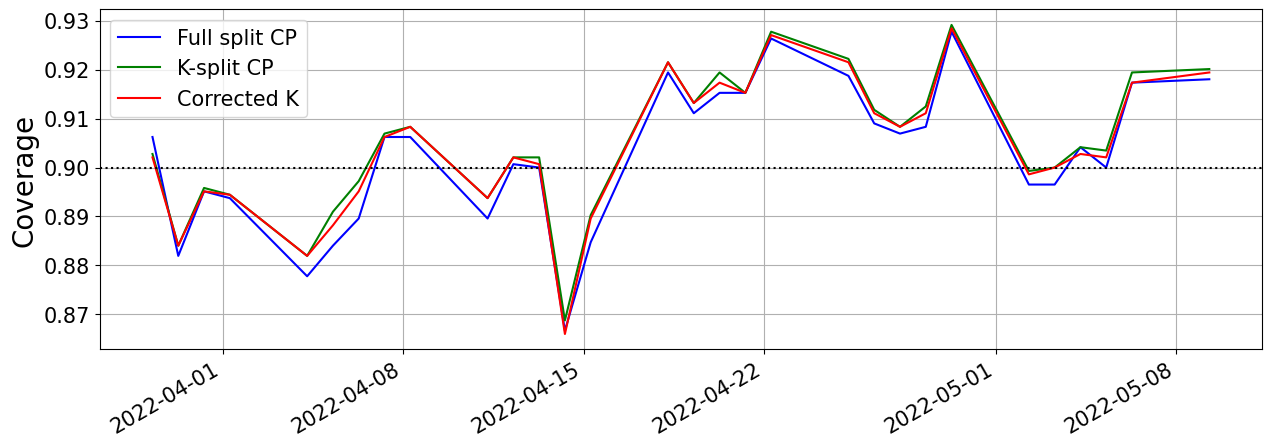}}
    \caption{Daily coverage for EUR/SEK exchange rate $(n=43200, N=43200)$ during one month}
    \label{fig:cov_eur_sek}
    \end{center}
\vskip -0.2in
\end{figure}

In Figure \ref{fig:cov_eur_sek}, we plotted the empirical averages of the coverage over a month. We observe that all three methods achieve $1-\alpha$ coverage. 

\subsubsection{Electricity price forecasting}

In this second example, we consider the same dataset as \cite{zaffran2022adaptive}, which contains the French electricity price between 2016 to 2019, reported every hour. We consider again the prediction of the one-step return $r_{t+1}$ with a rolling window of fixed size (18 months = 18x30x24 data points for both training/calibration). In this example, we obtained a value of $\rho=0.78$. We calculate the empirical coverage for the year 2019.

\begin{figure}[ht]
\vskip 0.2in
\begin{center}
\centerline{\includegraphics[width=\columnwidth]{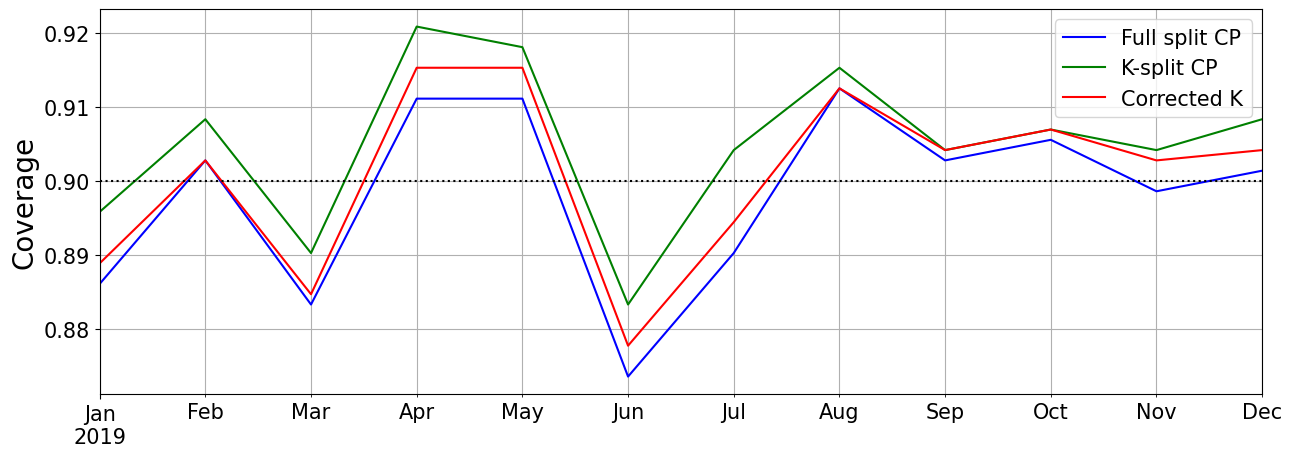}}
    \caption{Monthly coverage for French electricity price forecast $(n=12 960, N=12 960)$}
    \label{fig:elec}
    \end{center}
\vskip -0.2in
\end{figure}

In Figure \ref{fig:elec}, we can observe that all methods are quite similar in performance, with a coverage close to $1-\alpha$.

%% file: 11.conclusion.tex
\section{Conclusion}

In this paper, we extended the analysis of the  original split CP method to the case of Markovian data. We established upper bounds on the impact of the correlations in the data on the coverage guarantees and size of the prediction set. When the underlying Markov chain mixes rapidly, this impact is negligible. When this is not the case, handling correlations remains challenging and an interesting topic for future research. We could try for example to identify fundamental limits on the coverage vs. size of the prediction set trade-off satisfied by any conformal method; such limits would indicate the incompressible price one has to pay when dealing with Markovian data. These limits would also provide insights into the design of CP algorithms.  



%% file: 9.proofs.tex
\appendix
\section*{Appendix}

\section{Proofs}

\subsection{Notation}


We will often make use of the following definition of the total variation between two measures $P,Q$ defined on a measurable space $(\mathcal{X}, \mathcal{F})$ 

\begin{equation}
\label{dtv}
||P-Q||_{TV}=\sup_{A\in\mathcal{F}} |P(A)-Q(A)| 
\end{equation}

We will also make use of the equivalent formulation of 

$$\beta(a) =\sup_{t\ge 1} \| \mathbb{P}_1^t\otimes \mathbb{P}_{t+a}^\infty -\mathbb{P}_{t,a}\|_{TV}$$

for $a\ge 1$, 
where $\mathbb{P}_{t,a}$ is the joint distribution of $(X_1^t,X_{t+a}^\infty)$ \cite{doukhan2012mixing}.

It can be shown \cite{gallegos2023equivalences} that if a Markov chain is geometrically ergodic, then for any measure $\nu \in L^p(\pi)=\{\mu, \int |\frac{d\mu}{d\pi}|^p d\pi<\infty\}$ for some $p\ge 1$, there exists $0\le \rho<1$ (that depends on $\nu$ through $p$ only) and a constant $C_\nu$ such that for all $n\ge 1$, $\| \nu P^n-\pi\|_{TV}\le C_\nu\rho^n$. Without loss of generality, we will identify for conciseness the constants $C, C_{\nu}$ (resp $C'$) and $\rho$ appearing in the total variation term $\| \nu P^n-\pi\|_{TV}$ and in $\beta(a)$ (resp $\beta'(a)$) to be the same (consider the maximum of each value for example). 


We define the following probability notations: (i) the probability over the calibration dataset given the training dataset, $\mathbb{P}_\mathrm{cal}=\mathbb{P}_{\mathrm{cal},(X_{n+1},Y_{n+1})}(.|D_\mathrm{tr})$, (ii) the probability $\mathbb{P}_\pi$ when the test point $(X_{n+1},Y_{n+1})$ is taken independently of the chain and follows the stationary distribution $\pi$.

\subsection{Proof of Proposition \ref{prop:main} and Theorem \ref{thm:main}}

\begin{proof}

The proof is divided into 4 parts:

\begin{itemize}
    \item Calculation of the marginal coverage when the training and calibration datasets are independent
    \item Calculation of the marginal coverage in the general case
    \item Optimisation of the coverage gap w.r.t $u$
    \item Optimisation w.r.t $r$
\end{itemize}

Define the random variable $Z=(X,Y)$ and let $Z_{1-N}^{n+1}=((X_i,Y_i))_{i=1-N}^{n+1}$

\textbf{Calculation of the marginal coverage when the training and calibration datasets are independent}

    Suppose in this section only that $Z_{1-N}^{0}$ is fixed therefore $S$ is fixed and furthermore suppose that $Z_{1-N}^{0}$ is independent of $Z_1^{n}$
    
    Similar to \cite{foygel2021limits, oliveira2022split}, we show with high probability over $Z_1^{n}$ that $\hat{q}_{\alpha,n}$ is lower bounded by a true quantile $\hat{q}_{\alpha^+}$ of the scores $S(X,Y)$ when $(X,Y)\sim\pi$, and where $\alpha^+=\alpha+u>\alpha$ for $1-\alpha\geq u>0$.

Define the function $f$

$$f(Z_1^{n})=\sum_{i=1}^{n}\mathbbm{1}\{S_i> \hat{q}_{\alpha^+}\}$$

Then by definition on the calibration dataset
\begin{align}
\begin{split}
\label{eq:expectation}
    \mathbb{E}_\mathrm{cal}f(Z_1^{n})&=\sum_{i=1}^{n}\mathbb{P}_\mathrm{cal}(S_i> \hat{q}_{\alpha^+})\\
    &= \sum_{i=1}^{n}\mathbb{P}_\mathrm{cal}(S_i> \hat{q}_{\alpha^+})-\mathbb{P}_\pi(S> \hat{q}_{\alpha^+})+\mathbb{P}_\pi(S> \hat{q}_{\alpha^+}) \\
    &\geq\sum_{i=1}^{n}[\mathbb{P}_\pi(S> \hat{q}_{\alpha^+})-||\nu_0P^{N+i}-\pi||_{TV}] \\
    &= n\alpha^+ - \frac{1}{n}\sum_{i=1}^{n}||\nu_0P^{N+i}-\pi||_{TV}\\
    &\geq n\alpha^+ -\delta(N)
    \end{split}
\end{align}
By \cite{paulin2015concentration}, we have the following McDiarmid concentration

$$\forall t>0, \quad \mathbb{P}_\mathrm{cal}(\frac{1}{n}(f(Z_1^{n})-\mathbb{E}_\mathrm{cal}f(Z_1^{n}))> -t)\geq 1-\exp\bigg(\frac{-2n^2t^2}{||c||^2\tau_\textrm{min}}\bigg)$$

where

$$
\tau_\textrm{min}=\inf_{0\leq\varepsilon<1}\tau(\varepsilon)\big(\frac{2-\varepsilon}{1-\varepsilon}\big)^2 \qquad \textrm{and} \qquad \tau_\textrm{min}\leq 9t_\textrm{mix} 
$$

and $c\in \mathbb{R}^{n+1}=(1,...,1)^T, ||c||^2=n$  is a vector chosen such that for any $(Z,Z')$

$$f(Z_1^{n})-f(Z_1^{'n})\leq \sum_{i=1}^{T}c_i1\{Z_i\neq Z'_i\}$$

\bigskip
From (\ref{eq:expectation}), we have
\begin{align*}
    \frac{1}{n}(f(Z_1^{n})-\mathbb{E}_\mathrm{cal}f(Z_1^{n})) \geq - t \implies &\frac{1}{n} 
    \sum_{i=1}^{n} \mathbbm{1}\{S_i>\hat{q}_{\alpha^+}\} 
    \geq \alpha^+ -\delta(N)- t \\
    \implies &\sum_{i=1}^{n} \mathbbm{1}\{S_i>\hat{q}_{\alpha^+}\} 
    \geq n\alpha + n(u -\delta(N) - t) \\
    \implies &\sum_{i=1}^{n} \mathbbm{1}\{S_i>\hat{q}_{\alpha^+}\} 
    \geq n\alpha \quad \text{for $t = u -\delta(N)$}
\end{align*}
By definition of $\hat{q}_{\alpha,n}$, this inequality implies that $\hat{q}_{\alpha^+}\leq \hat{q}_{\alpha,n}$.

\bigskip

For any $1\leq r\leq n$ and $u>\delta(N)$, let $E_r(u)=\{f(Z_r^{n+1})\geq\mathbb{E}_\mathrm{cal}(f(Z_r^{n+1})) - (u -\delta(N))\}$. Therefore, for any $1\leq r\leq n$, we proved that under the hypothesis that $Z_{1-N}^{0}$ is independent of $Z_r^{n+1}$, then we have over both the training and (reduced) calibration datasets
\begin{equation}
\label{eq:quantile_ineq}
    \mathbb{P}(\hat{q}_{\alpha^+}\leq \hat{q}_{\alpha,(n-r)})\geq \mathbb{P}(E_r(u))\geq 1-\exp\bigg(\frac{-2(n-r)(u -\delta(N))^2}{9t_\textrm{mix}}\bigg)
\end{equation}

where $\hat{q}_{\alpha,(n-r)}$ is the empirical quantile over $\{S(Z_r^n)\}$.

\bigskip
Finally, we relate $\hat{q}_{\alpha,(n-r)}$ to $\hat{q}_{\alpha',n}$ with $\alpha'$ well chosen. 

For any dataset $D$ and point $Z\in D$, let $\mathrm{rank}(D,Z)$ be the rank of $Z$ in $D$. Then by definition of the empirical quantile, $\mathrm{rank}(D_{\mathrm{cal},r},\hat{q}_{\alpha,(n-r)}) = \lceil (n+r-1)(1-\alpha)\rceil$. When we add the remaining $S_1,..,S_r$ then $\mathrm{rank}(D_{\mathrm{cal}},\hat{q}_{\alpha,(n-r)}) \in [\lceil (n+r-1)(1-\alpha)\rceil-r, \lceil (n+r-1)(1-\alpha)\rceil+r]$. The extremities of the interval correspond to the cases where either $\min(S_1,...,S_r)\geq\max(S_{r+1},...,S_n)$ or $\max(S_1,...,S_r)\leq \min(S_{r+1},...,S_n)$. 

Hence, we directly obtain that 
$\hat{q}_{\alpha+\frac{1+r\alpha}{n+1},n}\leq \hat{q}_{\alpha,(n-r)} \leq \hat{q}_{\alpha-\frac{1+r\alpha}{n+1},n}$ and equivalently $\hat{q}_{\alpha+\frac{1+r\alpha}{n+1},(n-r)}\leq \hat{q}_{\alpha,n}\leq \hat{q}_{\alpha-\frac{1+r\alpha}{n+1},(n-r)}  $

\textbf{Calculation of the marginal coverage in the general case} 

We now consider the probability over the entire chain, therefore the training and calibration datasets are not independent. Similar to \cite{oliveira2022split}, for $r\geq 1$, we consider the reduced chain $Z_r^n$. Let $\mathbb{P}^\star_r$ be the joint distribution under which $Z_{1-N}^0$ and $Z_r^n$ are supposed independent. Let $\mathbb{P}^\star_{n+1}$ be the joint distribution under which $Z_{1-N}^0$ and $Z_{n+1}$ are supposed independent.

For any $u>\delta(N)$, for any $r\in[1,n]$, we have

\begin{align*}
    &\mathbb{P}(S_{n+1}\leq \hat{q}_{\alpha,n}) \\
    &\geq\mathbb{P}(S_{n+1}\leq \hat{q}_{\alpha',(n-r)})  \qquad \text{where $\alpha'=\alpha+\frac{1+r\alpha}{n+1}$}\\
    &\geq\mathbb{P}(S_{n+1}\leq \hat{q}_{\alpha',(n-r)},Z_r^n\in E_{r}(u)) \\
    &\geq\mathbb{P}(S_{n+1}\leq \hat{q}_{\alpha'+u},Z_r^n\in E_r(u) ) \qquad \text{since on $E_r(u)$, $\hat{q}_{\alpha'+u}\leq \hat{q}_{\alpha',n-r}$  and the c.d.f is non decreasing}\\
    &\geq \mathbb{P}(S_{n+1}\leq \hat{q}_{\alpha'+u}) - \mathbb{P}(Z_r^n\in E_r(u)) \\
    &\geq\mathbb{P}_{n+1}^\star(S_{n+1}\leq \hat{q}_{\alpha'+u}) - \beta(n+1) - \mathbb{P}_r^\star(Z_r^n\in E_r(u)) -\beta(r)\\
    & \geq\mathbb{P}_\pi(S\leq \hat{q}_{\alpha'+u}) + [\mathbb{P}_{n+1}^\star(S_{n+1}\leq \hat{q}_{\alpha'+u}) - \mathbb{P}_\pi(S\leq \hat{q}_{\alpha'+u})] -\beta(n+1) \\
    &- \exp\bigg(\frac{-2(n-r)(u-\delta(N))^2}{9t_\textrm{mix}}\bigg) - \beta(r) \\
    &\geq 1-\alpha-u - \frac{1+r\alpha}{n+1} - \delta(n+N+1) - \exp\bigg(\frac{-2(n-r)(u-\delta(N))^2}{9t_\textrm{mix}}\bigg) - \beta(n+1)-\beta(r)\\
    \end{align*}

where the fifth inequality comes from

\begin{align*}
    \mathbb{P}(S_{n+1}\leq \hat{q}_{\alpha'+u}) &= \mathbb{P}(S_{n+1}\leq \hat{q}_{\alpha'+u})  - \mathbb{P}_{n+1}^\star(S_{n+1}\leq \hat{q}_{\alpha'+u}) + \mathbb{P}_{n+1}^\star(S_{n+1}\leq \hat{q}_{\alpha'+u}) \\
    &\geq \mathbb{P}_{n+1}^\star(S_{n+1}\leq \hat{q}_{\alpha'+u}) - ||\mathbb{P}_{0, n+1} - \mathbb{P}_{1-N}^{0}\otimes  \mathbb{P}_{n+1}^{\infty}||_{TV} \\
    &\geq \mathbb{P}_{n+1}^\star(S_{n+1}\leq \hat{q}_{\alpha'+u}) - \sup_t||\mathbb{P}_{t, n+1} - \mathbb{P}_{1-N}^{t}\otimes  \mathbb{P}_{t+n+1}^{\infty}||_{TV}\\
    &= \mathbb{P}_{n+1}^\star(S_{n+1}\leq \hat{q}_{\alpha'+u}) -\beta(n+1)
\end{align*}

and similarly for $\mathbb{P}(Z_r^n\in E_r(u))$.

Notice that in the case of a restart of the chain at the calibration with initial distribution $\nu_1$, we can choose $r=0$ and ignore the $\beta$ terms therefore obtaining 

\begin{equation}
\label{eq:gamma_res}
    \gamma(u)=u 
 + \exp\bigg(\frac{-2n(u-\frac{1}{n}\sum_{i=1}^n\delta_1(i))^2}{9t_\textrm{mix}}\bigg) + \delta_1(n+1)
\end{equation}

If $X$ is $\phi$-irreductible aperiodic Markov chain with coefficients $\beta(a)$, we have 

\begin{equation}
    \label{eq:gamma_gen}
    \gamma(u,r) = u + \exp\bigg(\frac{-2(n-r)(u-\delta(N))^2}{9t_\textrm{mix}}\bigg) + + \delta(n+N+1) + \beta(n+1)+\beta(r)
\end{equation}

If $X$ is a $\phi$-irreductible aperiodic geometric ergodic Markov chain,

\begin{equation}
\label{eq:gamma_erg}
    \gamma(u,r)=u + \frac{1+r\alpha}{n+1}
 + \exp\bigg(\frac{-2(n-r)(u-C\rho^N)^2}{9t_\textrm{mix}}\bigg) + C(\rho^r+\rho^{n+1}+\rho^{n+N+1})
\end{equation}

\textbf{Upper bound}

We keep the same notations but replace $\alpha^+$ by $\alpha^-=\alpha+\frac{\alpha-1}{n}-u<\alpha$ where the additional $\frac{\alpha-1}{n}$ takes into account the correction in the empirical quantile calculation. 

Again, supposing that $Z_{1-N}^{0}$ is independent of $Z_1^{n+1}$ then, we have 

$$\forall t>0, \quad \mathbb{P}_\mathrm{cal}(\frac{1}{n}(f(Z)-\mathbb{E}_\mathrm{cal}f(Z))\leq t)\geq 1-\exp\bigg(\frac{-2nt^2}{9\tau_\textrm{min}}\bigg)$$

and we also have
\begin{align*}
    \frac{1}{n}(f(Z)-\mathbb{E}_\mathrm{cal}f(Z)) \leq t \implies &\frac{1}{n} 
    \sum_{i=1}^{n} \mathbbm{1}\{S_i>\hat{q}_{\alpha^-}\} 
    \leq \alpha^- + \delta(N) +  t \\
    \implies &\sum_{i=1}^{n} \mathbbm{1}\{S_i>\hat{q}_{\alpha^-}\} 
    \leq (n+1)\alpha -1 + n(t - (u-\delta(N))) \\
    \implies &n - \sum_{i=1}^{n} \mathbbm{1}\{S_i\leq \hat{q}_{\alpha^-}\} 
    \leq (n+1)\alpha -1 \quad \text{for $t = u-\delta(N)$} \\
    \implies &\sum_{i=1}^{n} \mathbbm{1}\{S_i\leq \hat{q}_{\alpha^-}\}>(n+1) (1-\alpha)\geq\lceil (n+1) (1-\alpha)\rceil \quad \text{since the l.h.s is an integer}
\end{align*}

By definition of $\hat{q}_{\alpha,n}$, this inequality implies that $\hat{q}_{\alpha^-}\geq \hat{q}_{\alpha,n}$. 

For any $1\leq r\leq n$ and $u>\delta(N)$, let $E_r(u)=\{f(Z_r^{n+1})\leq\mathbb{E}_\mathrm{cal}(f(Z_r^{n+1})) + (u-\delta(N))\}$.

\bigskip
We have 

\begin{align*}
    \mathbb{P}(S_{n+1}\leq \hat{q}_{\alpha,n}) &\leq \mathbb{P}(S_{n+1}\leq \hat{q}_{\alpha',(n-r)})
    \qquad \text{where $\alpha'=\alpha-\frac{1+r\alpha}{n+1}$}\\
    &=\mathbb{P}(S_{n+1}\leq \hat{q}_{\alpha',(n-r)},Z_r^n\in E_r(u)) + \mathbb{P}(S_{n+1}\leq \hat{q}_{\alpha',(n-r)},Z_r^n\in E_r(u)^c) \\
    &\leq \mathbb{P}(S_{n+1}\leq \hat{q}_{\alpha'-u},Z_r^n\in E_r(u)) + \mathbb{P}(Z_r^n\in E_r(u)^c) \\
    &\leq \mathbb{P}(S_{n+1}\leq \hat{q}_{\alpha'-u}) + \mathbb{P}(Z_r^n\in E_r(u)^c) \\
    &\leq\mathbb{P}_{n+1}^\star(S_{n+1}\leq \hat{q}_{\alpha'-u}) + \beta(n+1) + \mathbb{P}_r^\star(Z_r^n\in E_r(u)) +\beta(r) \\
    &\leq ...
\end{align*}

Hence, we obtain the same value of $\gamma$ with an additional $\frac{1-\alpha}{n-r}$ for all cases as in \ref{eq:gamma_res},\ref{eq:gamma_gen} and \ref{eq:gamma_erg}.

\bigskip

\textbf{Optimisation of $\gamma(u,r)$ w.r.t to $u$}

    We study the behaviour of $\gamma$ as a function of $u$.

    For simplicity, let $\gamma(u)=u+\exp(-A(u-K)^2)+B$ where 
    
    $$A=\frac{2(n-r)}{9t_\textrm{mix}} \quad K = \delta(N) \quad B = \frac{1+r\alpha}{n+1} +  \delta(n+N+1) + \beta(n+1)+\beta(r)$$

    We study the behaviour of the shifted function $u\rightarrow \gamma(u+K)$. 

By writing the first derivative, we have for $u\geq \delta(N)$

$$\gamma'(u)=1-2Au\exp(-Au^2)$$
Hence by equivalence (as $u>0$)
\begin{align*}
    \gamma'(u)=0 &\iff 1= 2A u\exp(-Au^2) \\
    &\iff 1 = 4A^2u^2\exp(-2 u^2) \\
    &\iff -\frac{1}{2A} = -2A u^2\exp(-2Au^2) 
\end{align*}
Let $w=-\frac{1}{2A}$ and $z=-2A u^2$ then we need to solve the following Lambert equation $ze^z=w$ with unknown $z$.

\bigskip
If $w<\frac{-1}{e}$ i.e $n-r< \frac{9e}{4}t_\textrm{mix}\equiv en_0$ then this equation has no solution as $\forall u, \quad ue^u>\frac{-1}{e}$. And by ascending the previous calculus by equivalence, this implies $\gamma'(u)>0$.

\bigskip
If $n-r\geq en_0$ then there exists two critical points $z_1=W_0(w)$ and $z_2=W_{-1}(w)$, which in turn gives 
$$u_1 = \sqrt{-W_0(-\frac{1}{2A})\frac{1}{2A}} \qquad u_2 =\sqrt{-W_{-1}(-\frac{1}{2A})\frac{1}{2A}} \qquad u_1\leq u_2$$ 
To know the ordering between both, we need to study the second order derivative.
$$\gamma''(u)=2C\exp(-A u^2)(2A u^2-1)$$
Hence, $\forall u<\sqrt{\frac{1}{2A}}=u_0, \quad \gamma''(u)<0$ and $\forall u\geq u_0, \quad \gamma''(u)\geq0$.

\bigskip
This implies that $\gamma'$ is decreasing on $[0,u_0]$ and increasing on $[u_0,1]$. Since $\gamma'(0)=1$, $\lim_{\infty} \gamma' = 1$, and $\gamma'(u_0)<0$ (which naturally comes from the condition $n-r\geq n_0$), then $\gamma'$ is positive on $[0,u_1]\cup[u_2,\infty]$, and negative on $[u_1,u_2]$.

\bigskip
This can be summed up in the following variation table. 

\bigskip
\begin{center}
\begin{tikzpicture}
   \tkzTabInit[espcl = 1.5]{$u$ / 1 , $\gamma''(u)$ / 1, $\gamma'(u)$ / 1, $\gamma'(u)$ / 1, $\gamma(u)$ / 1 }{$0$, $u_1$, $u_0$, $u_2$, $+\infty$}
   \tkzTabLine{, -, , -,z ,+, , +,}
   \tkzTabVar{+/ 1, R/,
            -/$<0$, R/, +/1}
   \tkzTabIma{1}{3}{2}{$0$}  
   \tkzTabIma{3}{5}{4}{$0$}  
   \tkzTabLine{, +,z , -, ,-, z, +,}
   \tkzTabVar{-/ 1, +/$\gamma(u_1)$, R/, -/$\gamma(u_2)$, +/$\infty$}
   
\end{tikzpicture}
           \end{center}
           
\bigskip
Therefore, we proved that $\gamma(u_2)\leq\gamma(u_1)$. Finally, we add the constants to obtain the optimal expression of $\gamma(u,r)$ which will again be denoted by $\gamma(r)$.

\begin{equation}
    \label{eq:opt_gamma_r}
    \gamma(r)=\sqrt{-W_{-1}(-\frac{n_0}{n-r})\frac{n_0}{n-r}} + \exp\bigg(\frac{1}{2}W_{-1}(-\frac{n_0}{n-r})\bigg) + \frac{1+r\alpha}{n+1} + 
     \delta(N) +  \delta(n+N+1) + \beta(n+1)+\beta(r)
\end{equation}

Remark that in the restart case, the optimisation remains the same and we obtain

\begin{equation}
    \label{eq:opt_gamma_restart}
    \gamma=\sqrt{-W_{-1}(-\frac{n_0}{n})\frac{n_0}{n}} + \exp\bigg(\frac{1}{2}W_{-1}(-\frac{n_0}{n})\bigg) +
     \frac{1}{n}\sum_{i=1}^n\delta_1(i) +  \delta_1(n+1)
\end{equation}

Let us consider a few important points in the optimisation of $\gamma(r)$:

\begin{itemize}
    \item From \ref{eq:gamma_gen} or \ref{eq:gamma_erg} it is clear that for any $u$, $r\rightarrow\gamma(u,r)$ is non increasing.  Therefore it must also hold for the optimal value of $u=u_2$
    \item Hence $\sqrt{-W_{-1}(-\frac{n_0}{n-r})\frac{n_0}{n-r}} + \exp\bigg(\frac{1}{2}W_{-1}(-\frac{n_0}{n-r})\bigg) \geq \sqrt{-W_{-1}(-\frac{n_0}{n})\frac{n_0}{n}} + \exp\bigg(\frac{1}{2}W_{-1}(-\frac{n_0}{n})\bigg)=\gamma_\mathrm{as}$
    \item For our method to achieve optimal lower bound $\gamma_\mathrm{as}$, we are interested in processes with $\beta$ coefficient such that for a well chosen $r(n)=o_n(n)$ we have $\beta(r)={\cal O}_n(\gamma_\mathrm{as})$ and $\frac{r}{n}={\cal O}_n(\gamma_\mathrm{as})$
    \item Finally, this will yield an optimal asymptotic value (in the geometric ergodic case) $\gamma=\gamma(r(n))={\cal O}_{n,N}(\gamma_\mathrm{as})$
\end{itemize}

We start by analysing the asymptotic order of $\gamma_\mathrm{as}$. Using the inequality from \cite{chatzigeorgiou2013bounds}, we have

$$\forall z>0 \quad -1-\sqrt{2z}- z \leq W_{-1}(-e^{-z-1})\leq -1-\sqrt{2z}-\frac{2}{3} z$$

Hence, with $z=\ln\frac{n}{n_0}$, we have

$$\gamma_{\textrm{min}}\leq \gamma_\mathrm{as}\leq \gamma_{\textrm{max}}$$

where 
\begin{align*}
    \gamma_{\textrm{min}} &= \sqrt{\frac{n_0}{n}(1+\sqrt{2\ln\frac{n}{n_0}}+\frac{2}{3}\ln\frac{n}{n_0})}+\sqrt{\frac{n_0}{n}}\exp(-\sqrt{\frac{1}{2}\ln\frac{n}{n_0}}) \\
\gamma_{\textrm{max}}&=\sqrt{\frac{n_0}{n}(1+\sqrt{2\ln\frac{n}{n_0}}+\ln\frac{n}{n_0})}+\sqrt{\frac{1}{e}(\frac{n_0}{n})^{2/3}}\exp(-\sqrt{\frac{1}{2}\ln\frac{n}{n_0}})
\end{align*}

It is clear that $\gamma_{\textrm{min}}={\cal O}_n(\sqrt{\frac{n_0}{n}(1+\sqrt{2\ln\frac{n}{n_0}}+\frac{2}{3}\ln\frac{n}{n_0})})={\cal O}_n(\sqrt{\frac{n_0}{n}\ln\frac{n}{n_0}})$ 

\bigskip
For $\gamma_{\textrm{max}}$, we prove that $\frac{1}{n^{1/3}}\exp(-\sqrt{\frac{1}{2}\ln n})<\sqrt{\frac{\ln n}{n}}={\cal O}_n(\sqrt{\frac{n_0}{n}\ln\frac{n}{n_0}})$. This is equivalent to proving that for $n$ large

$$h(n)=\frac{1}{3}\ln n -\sqrt{2\ln n}-\ln\ln n<0$$

We have for any $z$ large enough

$$h'(z)=\frac{1}{z}(\frac{1}{3}-\frac{1}{\ln z}-\frac{1}{\sqrt{2\ln z}})$$
and the term inside the parenthesis can be seen as a polynomial of degree 2 with unknown $\frac{1}{\sqrt{\ln z}}$. The discriminant is strictly positive hence $h'$ is the sign of the leading coefficient which is $-1$ here. Therefore $h$ is decreasing and for $h(e)=\frac{1}{3}-\sqrt{2}<0$ hence $h$ is negative. 

\bigskip
This proves that $\gamma_{\textrm{max}}={\cal O}_n(\sqrt{\frac{n_0}{n}\ln\frac{n}{n_0}})$ and therefore $\gamma_\mathrm{as}$ has also the same rate of convergence i.e $\gamma_\mathrm{as}={\cal O}_n(\sqrt{\frac{n_0}{n}\ln\frac{n}{n_0}})={\cal O}_n (\sqrt{t_\mathrm{mix}{\ln(n)\over n}})$.

Hence, ignoring the total variation terms, we are looking for stochastic processes such that there exists $r=o_n(n)$ and $\beta(r)={\cal O}_n(\sqrt{t_\mathrm{mix}{\ln(n)\over n}})$.

\textbf{Optimisation of $\gamma(r)$ for geometrically ergodic chains}

We show that all conditions are fulfilled for geometric ergodic Markov chains and we can achieve asymptotically optimal rate of convergence $\gamma_\mathrm{as}$

Indeed, since $\beta(r)\leq C\rho^r$, taking $r=\frac{\ln n}{\ln\frac{1}{\rho}}$ gives $r=o_n(n), \frac{r}{n}={\cal O}_n(\gamma_\mathrm{as}), \beta(r)={\cal O}_n(\frac{1}{n})={\cal O}_n(\gamma_\mathrm{as})$ and the proof is complete.






 \end{proof}

\subsection{Proof of Proposition \ref{prop:length}}

\begin{proof}

\textbf{Comparison with $\mu$}

Let $q_{\alpha,n}$ be the empirical quantile of $\{|Y_i-\mu(X_i)|\}_{i=1}^n$.

Let $E=\{||\hat{\mu}_N- {\mu}||_\infty \leq c_N\}$ such that $\mathbb{P}(E)\geq 1-d_N$. 

We have

\begin{align*}
    |\hat{q}_{\alpha,n}-\hat{q}_\alpha|\leq |\hat{q}_{\alpha,n}- {q}_{\alpha,n}| + | {q}_{\alpha,n}- {q}_\alpha| + | {q}_\alpha -\hat{q}_\alpha|
\end{align*}

Following the proof of \cite{lei2018distribution}, we have for any $D_\mathrm{tr}\in E$,

$$| {q}_\alpha -\hat{q}_\alpha|\leq \frac{d_N}{\kappa}+c_N$$ 

and for any $D_\mathrm{cal}$

$$|\hat{q}_{\alpha,n}- {q}_{\alpha,n}|\leq c_N$$

Hence for any $u>2\kappa c_N+d_N$, we have 

$$\{| {q}_{\alpha,n}- {q}_\alpha|\leq \frac{u}{\kappa}-2c_N-\frac{d_N}{\kappa}\} \in \{|\hat{q}_{\alpha,n}-\hat{q}_\alpha|\leq \frac{u}{\kappa}\}$$

This gives
\begin{align*}
    \mathbb{P}(|\hat{q}_{\alpha,n}-\hat{q}_\alpha|\leq \frac{u}{\kappa}) &\geq \mathbb{P}(|\hat{q}_{\alpha,n}-\hat{q}_\alpha|\leq \frac{u}{\kappa}, E) \\
    &\geq \mathbb{P}(| {q}_{\alpha,n}- {q}_\alpha|\leq \frac{u}{\kappa}-2c_N-\frac{d_N}{\kappa}, E) 
\end{align*}

In the restart version, we can directly apply Equation \ref{eq:quantile_ineq} provided that $u'=u-2\kappa c_N -d_N \in \mathcal{N}$ and obtain that

\begin{align*}
    \mathbb{P}(|\hat{q}_{\alpha,n}-\hat{q}_\alpha|\leq \frac{u}{\kappa}) &\geq (1-2\exp\big[-\frac{2n}{9t_\textrm{mix}}\big(u'-\delta_1(N)\big)^2\big])\mathbb{P}_\mathrm{tr}(E)\\
    &\geq 1-2\exp\big[-\frac{2n}{9t_\textrm{mix}}\big(u'-\delta_1(N)\big)^2\big]-d_N
\end{align*}

\textbf{Application of the concentration inequality over $D_\mathrm{cal}$ as in Proposition $\ref{prop:main}$}

We now compute $\mathbb{P}(| {q}_{\alpha,n}- {q}_\alpha|\leq \frac{u}{\kappa}-2c_N-\frac{d_N}{\kappa}, E)$.

Since the concentration inequality was applied on the last $n-r$ samples, we need to replace $| {q}_{\alpha,n}- {q}_\alpha|$ by $| {q}_{\alpha',n-r}- {q}_\alpha|$ for a certain $\alpha'$. 

For any $r\in[1,n]$, we have

$${q}_{\alpha+d\alpha,(n-r)}\leq  {q}_{\alpha,n}\leq  {q}_{\alpha-d\alpha,(n-r)}$$

where $d\alpha=\frac{1+r\alpha}{n+1}$. Hence, we have for $v=\frac{u}{\kappa}-2c_N-\frac{d_N}{\kappa}$

$$q_\alpha - v \leq q_{\alpha,n}\leq q_\alpha +  v \implies q_\alpha+v\geq {q}_{\alpha+d\alpha,(n-r)} \quad , \quad q_\alpha-v\leq {q}_{\alpha-d\alpha,(n-r)}  $$


Supposing $d\alpha\in\mathcal{N}$, we also have by the bounded hypothesis on the density $ {f}$,  

$$(1-\alpha+d\alpha)-(1-\alpha) = \int_{ {q}_\alpha}^{ {q}_{\alpha-d\alpha}}  {f}\geq ( {q}_{\alpha-d\alpha}- {q}_\alpha)\kappa$$ hence 

$$ {q}_{\alpha-d\alpha}\leq  {q}_\alpha + \frac{d\alpha}{\kappa} \qquad \text{and} \quad  {q}_{\alpha+d\alpha}\geq  {q}_\alpha - \frac{d\alpha}{\kappa}$$

therefore

$$q_\alpha+v\geq {q}_{\alpha+d\alpha,(n-r)} \quad , \quad q_{\alpha}-v\leq {q}_{\alpha-d\alpha,(n-r)} \implies q_{\alpha+d\alpha} + \frac{d\alpha}{\kappa} + v \geq {q}_{\alpha+d\alpha,(n-r)}  \quad , \quad q_{\alpha-d\alpha}-\frac{d\alpha}{\kappa}-v\leq {q}_{\alpha-d\alpha,(n-r)}  $$

Finally, we know from Equation $\ref{eq:quantile_ineq}$ that for any $u'>\delta(N)$, we have provided that $u'\leq \alpha+d\alpha, u'\in \mathcal{N}$

$$ q_{\alpha+d\alpha,(n-r)}\leq q_{\alpha+d\alpha-u'} \leq q_{\alpha+d\alpha} + \frac{u'}{\kappa}$$

with probability at least $ 1-\exp\bigg(\frac{-2(n-r)(u' -\delta(N))^2}{9t_\textrm{mix}}\bigg)$ if the (reduced) calibration and training datasets are independent

and similarly

$$ q_{\alpha-d\alpha,(n-r)}\geq q_{\alpha-d\alpha+u'} \geq q_{\alpha-d\alpha} - \frac{u'}{\kappa}$$

Therefore identifying $\frac{u'}{\kappa}=\frac{d\alpha}{\kappa}+v$ i.e $u'=d\alpha +v\kappa$ we have 

\begin{align*}
    &\mathbb{P}(| {q}_{\alpha,n}- {q}_\alpha|\leq \frac{u}{\kappa}-2c_N-\frac{d_N}{\kappa}, E) \\
    &\geq  \mathbb{P}(  {q}_{\alpha-d\alpha}-{q}_{\alpha-d\alpha,n-r}\leq \frac{u-2\kappa c_N -d_N + d\alpha}{\kappa},{q}_{\alpha+d\alpha,n-r}-{q}_{\alpha+d\alpha}\leq \frac{u-2\kappa c_N -d_N + d\alpha}{\kappa}, E) \\
    &\geq  \mathbb{P}^\star_r(  {q}_{\alpha-d\alpha}-{q}_{\alpha-d\alpha,n-r}\leq \frac{u-2\kappa c_N -d_N + d\alpha}{\kappa},{q}_{\alpha+d\alpha,n-r}-{q}_{\alpha+d\alpha}\leq \frac{u-2\kappa c_N -d_N + d\alpha}{\kappa}, E)- \beta(r) \\
    &\geq (1-2\exp\big[-\frac{2(n-r)}{9t_\textrm{mix}}\big(u'-\delta(N)\big)^2\big])\mathbb{P}_\mathrm{tr}(E) -\beta(r) \\
    &\geq 1-2\exp\big[-\frac{2(n-r)}{9t_\textrm{mix}}\big(u'-\delta(N)\big)^2\big] - d_N -\beta(r)
\end{align*}

\end{proof}

\subsection{Proof of Proposition \ref{opt:length}}

The proof in \cite{lei2018distribution} consists in proving the two following points:
\begin{itemize}
    \item $\forall \varepsilon>0 \quad \lim_{N\rightarrow\infty} \mathbb{P}_{X_{n+1}, \mathrm{tr}}(|\hat{\mu}_N(X_{n+1})-\mu(X_{n+1})|>\varepsilon) = 0$ 
    \item $\forall \varepsilon>0 \quad \lim_{N\rightarrow\infty}\lim_{n\rightarrow\infty} \mathbb{P}(|\hat{q}_{\alpha,n}-q_\alpha|>\varepsilon) = \lim_{n\rightarrow\infty}\lim_{N\rightarrow\infty} \mathbb{P}(|\hat{q}_{\alpha,n}-q_\alpha|>\varepsilon)= 0$
    \end{itemize}

We tackle 1) the additional difficulty that $N$ and $n$ are different, 2) the non iid settings.

\begin{itemize}
    \item 
For the second point, we know by \cite{lei2018distribution} that $|\hat{q}_\alpha - q_\alpha|\le (M/2\kappa) \Delta_N$ hence

$$a_{N,n}=\mathbb{P}(|\hat{q}_{\alpha,n}-q_\alpha|>\varepsilon) \leq \mathbb{P}(|\hat{q}_{\alpha,n}-\hat{q}_\alpha|+ (M/2\kappa) \Delta_N>\varepsilon)$$

By the hypothesis that $\Delta_N=o_N(1), c_N= o_N(1), d_N= o_N(1)$, 
\begin{itemize}
    \item Let $N_0$ such that $\forall N\geq N_0 , \Delta_N\leq \varepsilon$
    \item Let $N_1\geq N_0$ such that $\forall N\geq N_1$, $f(N)=(M/2\kappa) \Delta_N + 2\kappa c_{N} +d_{N} \leq \kappa\varepsilon/2$.

\end{itemize}

\bigskip

Let $n_0$ such that $\forall n\geq n_0$, $\kappa\epsilon - \frac{1+\alpha r}{n+1} \geq f(N_1)$. This is justified as the l.h.s tends to $\kappa\varepsilon$ when $n\to\infty$ therefore it must be larger than $\kappa\varepsilon/2$ after an integer $n_0$.

Then by applying Proposition \ref{prop:length}  we have for any $n\geq n_0$,

$$\mathbb{P}(|\hat{q}_{\alpha,n}-q_\alpha|>\varepsilon)\leq 2e^{-\frac{2(n-\ln n)}{9t_\mathrm{mix}}\big(u'-\delta(N_1)\big)^2} +d_{N_0}+\beta(\ln n) $$

where $u'=\kappa\epsilon -(M/2\kappa) \Delta_{N_1} - 2\kappa c_{N_1} -d_{N_1} - \frac{1+\alpha \ln n}{n+1}$.

\bigskip

Let $N_2\geq N_1$ such that $\forall N\geq N_2$, $f(N)\leq f(N_1)$ 

Then $\forall n\geq n_0$ and $\forall N\geq N_2$  

$$\mathbb{P}(|\hat{q}_{\alpha,n}-q_\alpha|>\varepsilon)\leq 2e^{-\frac{2(n-\ln n)}{9t_\mathrm{mix}}\big(u'-\delta(N)\big)^2} +d_{N}+\beta(\ln n) $$

where $u'=\kappa\epsilon -(M/2\kappa) \Delta_N - 2\kappa c_{N} -d_{N} - \frac{1+\alpha \ln n}{n+1}$.

\bigskip

Now that we decoupled $n$ and $N$ (since all integers defined beforehand only depend on $\varepsilon$), we can take the limit w.r.t $n$ and $N$. Since it is a sum of terms that tend to 0 (supposing the chain is $\beta$-mixing) then the order does not count and we have 

$$\forall \varepsilon>0 \quad \lim_{N\rightarrow\infty}\lim_{n\rightarrow\infty} \mathbb{P}(|\hat{q}_{\alpha,n}-q_\alpha|>\varepsilon) = \lim_{n\rightarrow\infty}\lim_{N\rightarrow\infty} \mathbb{P}(|\hat{q}_{\alpha,n}-q_\alpha|>\varepsilon)= 0$$

\bigskip
\item To prove the first point, remark that the condition $\mathbb{E}_{X\sim \pi,\mathrm{tr}}[(\hat{\mu}_N(X)-\mu(X))^2]=o_N(1)$ implies that there exists sequences $\eta_N=o(1)$ and $\rho_N=o(1)$ such that

\begin{equation}
\label{eq:weak_consistence}
    \mathbb{P}(\mathbb{E}_{X\sim \pi}[(\hat{\mu}_N(X)-\mu(X))^2|\hat{\mu}_N]\geq \eta_N)\leq \rho_N
\end{equation}

Let $I=\{\Delta_N\leq \eta_N\}$ be the event which has probability at least $1-\rho_N$.

\bigskip

We calculate 

\begin{align*}
    \mathbb{P}(|\hat{\mu}_N(X_{n+1})-\mu(X_{n+1})|\geq \eta^{1/3})&\leq\mathbb{P}^\star_{n+1}(|\hat{\mu}_N(X_{n+1})-\mu(X_{n+1})|\geq \eta^{1/3}) + \beta(n+1) \\
    &= \mathbb{E}(\mathbb{P}^\star_{n+1}(|\hat{\mu}_N(X_{n+1})-\mu(X_{n+1})|\geq \eta^{1/3}|D_\mathrm{tr})) + \beta(n+1)
\end{align*}

Therefore by Markov inequality conditioned on $I$

$$\mathbb{P}^\star_{n+1}(|\hat{\mu}_N(X_{n+1})-\mu(X_{n+1})|\geq \eta_N^{1/3}|D_\mathrm{tr})\leq \frac{\mathbb{E}_{X_{n+1}^\star}(|\hat{\mu}_N(X_{n+1}^\star)-\mu(X_{n+1}^\star)|^2)}{\eta_N^{2/3}} $$

If $X_{n+1}\sim\pi$ then the r.h.s becomes $\eta_N^{1/3}$ and under $\beta$ mixing assumption, we prove the first point as we have

$$\mathbb{P}(|\hat{\mu}_N(X_{n+1})-\mu(X_{n+1})|\geq \eta^{1/3})\leq \eta_N^{1/3} + \rho_N + \beta(n+1) $$

and the r.h.s tends to 0 independently of the order for $n$ and $N$. 

\bigskip

Otherwise under $\mathbb{P}^\star$, $X_{n+1}\sim\nu_0P^{n+N+1}$ (or $\nu_1P^n$ in the restart version)

Suppose that we have furthermore Assumption \ref{assumption:stability} and let $I=\{\Delta_N\leq \eta_N, ||\hat{\mu}_N-\mu||_\infty\leq c_N\}$ be the event which has probability at least $1-\rho_N-d_N$. From the same calculus as above, we bound the r.h.s

\begin{align*}
    \mathbb{E}_{X_{n+1}^\star}(|\hat{\mu}_N(X_{n+1}^\star)-\mu(X_{n+1}^\star)|^2)&=\int_x (\hat{\mu}_N(x)-\mu(x))^2\nu_0P^{n+N+1}(dx) \\
    &\leq c_N^2
\end{align*}

Hence, we obtain 

$$\mathbb{P}(|\hat{\mu}_N(X_{n+1})-\mu(X_{n+1})|\geq \eta_N^{1/3})\leq  \frac{c_N^2}{\eta_N^{2/3}}+ \rho_N+d_N + \beta(n+1) $$

Furthermore, remark that Assumption \ref{assumption:stability} also implies Equation \ref{eq:weak_consistence} as

$$\mathbb{E}_{X\sim \pi}[(\hat{\mu}_N(X)-\mu(X))^2|\hat{\mu}_N]\leq ||\hat{\mu}-\mu||_\infty\leq c_N^2$$

where the last inequality holds with probability at least $1-d_N$ therefore we can take $\eta_N=c_N^2$ and $\rho_N=d_N$. Hence, we can conclude again by taking the limit, independently of the order, when $n,N\to\infty$. 

\end{itemize}

\subsection{Proof of Proposition \ref{proposition:cov_k_split}}

\begin{proof}
    For simplicity, we suppose that $K$ divides $n$ (otherwise the difference is asymptotically negligible). For any $r,s$, let $Z_{r}^s=\{(X_i,Y_i)\}_{i=r}^s$.

\textbf{Stationary case:}

We describe here the original blocking technique for \textit{stationary} $\beta$-mixing sequences with fixed block size.

\begin{itemize}
    \item Divide a sample $Z = (Z_1,...,Z_n)$ into $2m$ blocks $B_1,...,B_{2m}$. Even blocks are of size $a$ and odd blocks are of size $b$ i.e $n=2m(a+b)$ (suppose $n$ is even as there are minor changes otherwise)
    \item Denote by $B_\mathrm{odd}=(B_1, B_3,...,B_{2m-1})$ the list of odd blocks and define its independent version $B_\mathrm{odd}^\star=(B_1^\star, B_3^\star,...,B_{2m-1}^\star)$ where 
    
    $$B_{2k-1}^\star\stackrel{d}{=}B_{2k-1} \qquad B_1^\star \perp B_3^\star...$$ 
\end{itemize}

Then we have the following theorem

\begin{theorem}[Blocking technique \cite{yu1994rates}]
\label{thm:block}
    For all bounded measurable functions $|h|\leq M$

$$|\mathbb{E}(h(B_\mathrm{odd}))-\mathbb{E}(h(B_\mathrm{odd}^\star))|\leq 2Mm\beta(a)$$

\end{theorem} 

\bigskip
1) Suppose in the following only that $Z_{1-N}^0$ and $Z_1^n$ are independent and let $Z_{1-N}^{0}$ fixed such that $S$ is well defined and constant.

Let $B_\mathrm{odd}=\{Z_{iK+1}\}_{i=0}^{\frac{n}{K}}= \{Z_{iK+1}\}_{i=0}^{\frac{n}{K}-1}\cup Z_{n+1}$ and $h(B_\mathrm{odd})=1\{S_{n+1}\leq \hat{q}_{\alpha,K,n}\}$. 

Conditionally on $Z_{1-N}^0$, $Z_1^n$ still follows the stationary distribution as they are independent. Therefore applying Theorem \ref{thm:block}, we obtain
    
$$ |\mathbb{P}_{\textrm{cal},Z_{n+1}}(S_{n+1}\leq \hat{q}_{\alpha,K,n})-\mathbb{P}_{\textrm{cal}^\star,Z_{n+1}^\star}(S_{n+1}\leq \hat{q}_{\alpha,K,n})|\leq2\frac{n}{K}\beta_\mathrm{cal}(K) $$

where under $\mathbb{P}_{\textrm{cal}^\star,Z_{n+1}^\star}$, the samples $\{Z_{iK+1}\}_{i=0,...,\frac{n}{K}}$ are i.i.d. Since the calibration dataset is independent of the training dataset then $\beta_\mathrm{cal}(K)$ can be seen simply as the original $\beta(K)$ coefficient but on a smaller chain. Therefore, $\beta_\mathrm{cal}(K)\leq\beta(K)$.

It is well known that

$$1-\alpha\leq \mathbb{P}_{\textrm{cal}^\star,Z_{n+1}^\star}(S_{n+1}\leq \hat{q}_{\alpha,K,n}) \leq 1-\alpha+ \frac{1}{1+\frac{n}{K}}$$

From there, we can conclude for the restart part in the stationary case.

\bigskip
2) Suppose now that $Z_{1-N}^0$ and $Z_1^n$ are no longer independent. Denote by $\mathbb{P}^\star_{r,\textrm{cal},Z_{n+1}}$ the conditional probability under which $Z_r^{n+1}$ is independent of $Z_{1-N}^0$. Similarly, denote by $\mathbb{P}^\star_{r,\textrm{cal}^\star,Z_{n+1}^\star}$ the same probability and with $\{Z_i\}_{i=r}^{n+1}$ i.i.d.

\begin{align*}
    \mathbb{P}(S_{n+1}\leq \hat{q}_{\alpha,K,n}) &\geq  \mathbb{P}(S_{n+1}\leq \hat{q}_{\alpha',K,n-r}) \quad \text{where $\alpha'=\alpha+\frac{1+\alpha r}{n+1}$} \\
    &\geq \mathbb{P}_r^\star(S_{n+1}\leq \hat{q}_{\alpha',K,n-r}) -\beta(r) \\
    &= \mathbb{E}[\mathbb{P}_{r, \textrm{cal}, Z_{n+1}}^\star(S_{n+1}\leq \hat{q}_{\alpha',K,n-r})] -\beta(r) \\
    &\geq \mathbb{E}[\mathbb{P}^\star_{r,\textrm{cal}^\star,Z_{n+1}^*}(S_{n+1}\leq \hat{q}_{\alpha',K,n-r}) ]- 2\frac{n-r}{K}\beta(K) - \beta(r) \\
    &= 1-\alpha - \frac{1+\alpha r}{n+1} - 2\frac{n-r}{K}\beta(K) - \beta(r)
\end{align*}

and similarly, 

$$\mathbb{P}(S_{n+1}\leq \hat{q}_{\alpha,K,n-r})\leq 1-\alpha + \frac{K}{n-r}+\frac{1+\alpha r}{n+1}+2\frac{n-r}{K}\beta(K) + \beta(r) $$

\textbf{Non stationary case}

1) Suppose again in the following only that $Z_{1-N}^0$ and $Z_1^n$ are independent and let $Z_1^n$ fixed such that $S$ is well defined and constant. However, $Z_{1}^n$ is not necessarily stationary anymore as for all $i\in [1,n], \mathbb{P}_i=\mathbb{P}(Z_i)=\nu_0P^{N+i}$.

We will use an extended result of the original "Blocking Technique" for \textit{non stationary} $\beta$-mixing sequences which can be found in \cite{kuznetsov2017generalization}. More precisely, suppose $n=ma$ for $a\geq 1, m\geq 1$, and define for $j=0,
...,a-1$, the list $B_\mathrm{odd}^j=(Z_{1+j}, Z_{1+a+j},...,Z_{1+(m-1)a+j})$ where each sample are separated by $a$. Then we have the following proposition

\bigskip

\begin{proposition}[Proposition 2 of \cite{kuznetsov2017generalization}]
\label{prop:blocking_technique}

For all bounded measurable functions $|h|\leq M$ we have 

$$|\mathbb{E}(h(B_\mathrm{odd}^j))-\mathbb{E}(h(B^\star_\pi))|\leq 2Mm\beta'(a)$$

where $\beta'(a)=\sup_{t\geq1}\mathbb{E}(||P^{t+a}(.|(Z_1,...,Z_t))-\pi||_{TV})$ and $B^\star_\pi$ is an i.i.d sample of size $m$ which follows the stationary distribution $\pi$.
\end{proposition}

Let $B_\mathrm{odd}=\{Z_{iK+1}\}_{i=0}^{\frac{n}{K}}= \{Z_{iK+1}\}_{i=0}^{\frac{n}{K}-1}\cup Z_{n+1}$ and $h(B_\mathrm{odd})=1\{S_{n+1}\leq \hat{q}_{\alpha,K,n}\}$. Then applying Proposition \ref{prop:blocking_technique} to $B_\mathrm{odd}$, we obtain 
    
$$ |\mathbb{P}_{\textrm{cal},Z_{n+1}}(S_{n+1}\leq \hat{q}_{\alpha,K,n})-\mathbb{P}_{\textrm{cal}^\star,Z_{n+1}^\star}(S_{n+1}\leq \hat{q}_{\alpha,K,n})|\leq2\frac{n}{K}\beta'_{1}(K) $$

where $\forall i,  0\leq i\leq \frac{n}{K}$, $Z_{iK+1}^\star$ are independent and follow the stationary distribution $\pi$.

Therefore, again we fall back under the i.i.d settings and it is well-known that 

$$1-\alpha\leq \mathbb{P}_{\textrm{cal}^\star,Z_{n+1}^\star}(S_{n+1}\leq \hat{q}_{\alpha,K,n}) \leq 1-\alpha+ \frac{1}{1+\frac{n}{K}}$$

We can conclude for the restart version in the non stationary case.

2) Suppose now that $Z_{1-N}^0$ and $Z_1^n$ are no longer independent then 

\begin{align*}
\mathbb{P}(S_{n+1}\leq \hat{q}_{\alpha,K,n}) &\geq
    \mathbb{P}(S_{n+1}\leq \hat{q}_{\alpha',K,n-r}) \qquad \text{where $\alpha'=\alpha+\frac{1+\alpha r}{n+1}$} \\
    &\geq \mathbb{P}_r^\star(S_{n+1}\leq \hat{q}_{\alpha',K,n-r}) -\beta(r) \\
    &= \mathbb{E}[\mathbb{P}_{r, \textrm{cal}, Z_{n+1}}^\star(S_{n+1}\leq \hat{q}_{\alpha',K,n-r})] -\beta(r) \\
    &\geq \mathbb{E}[\mathbb{P}^\star_{r,\mathrm{cal}^\star,Z_{n+1}^*}(S_{n+1}\leq \hat{q}_{\alpha,K,n-r}) ]- 2\frac{n-r}{K}\beta'_{r}(K) - \beta(r) \\
    &= 1-\alpha - \frac{1+\alpha r}{n+1} - 2\frac{n-r}{K}\beta'_{r}(K) - \beta(r)
\end{align*}

and similarly, 

$$\mathbb{P}(S_{n+1}\leq \hat{q}_{\alpha,K,n})\leq 1-\alpha + \frac{K}{n-r}+\frac{1+\alpha r}{n+1}+2\frac{n-r}{K}\beta'_{r}(K) + \beta(r) $$

\end{proof}

\subsection{Proof of Theorem \ref{thm:cov_k}}

We focus our analysis on the "without restart" version as the "restart" version is simply an application of the former.

    We suppose $K>>1$ (which will be verified further on) and we wish to minimize the total coverage gap as a function of $K$ (i.e without the terms dependeing on $r$ only)
    $$\gamma_K=\frac{K}{n-r}+a\frac{n-r}{K}\rho^{K}$$

    (replace $\rho$ by $\sqrt{\rho}$ in the non stationary case)

    For $u>0$, let $f(u)=\frac{u}{n}+a\frac{n}{u}\rho^{u}$ then

    \begin{align*} 
        &f'(u)=0 \\
        &\iff \frac{1}{n-r}+a\frac{n-r}{u}e^{u\ln \rho}(\ln \rho -\frac{1}{u})=0 \\
        &\iff \frac{u^2}{n-r}+a(n-r) e^{u\ln \rho}(u\ln\rho-1)=0
    \end{align*}

Suppose furthermore that the optimal $K$ (and therefore $u$) verifies $|K\ln\rho|>>1$ then the last equation is asymptotically equal to 

\begin{align*}
&\frac{u^2}{n-r}+a(n-r) u\ln(\rho) e^{u\ln \rho}=0 \\
&\iff
-u\ln(\rho) e^{-u\ln\rho}=a(n-r)^2(\ln\rho)^2\\
&\iff u=\frac{W_0(a(n-r)^2(\ln \rho)^2)}{\ln\frac{1}{\rho}}={\cal O}_n  (\frac{\ln (a(n-r)^2(\ln \rho)^2)}{\ln \frac{1}{\rho}})={\cal O}_n (\frac{\ln (n-r)}{\ln \frac{1}{\rho}})
\end{align*}

Let us again consider a few important points when choosing the optimal value of $r$: 

\begin{itemize}
    \item The coverage gap is an increasing function of $r$ 
    \item When $r=0$, $K=\frac{W_0(an^2(\ln \rho)^2)}{\ln\frac{1}{\rho}}$ (as it indeed verifies $K>>1$ and $|K\ln\rho|>>1$) therefore 
    
    $$\frac{K}{n}={\cal O}_n(\frac{\ln n}{n\ln\frac{1}{\rho}})$$
    and
    \begin{align*}
        a\frac{n}{K}\rho^K &= {\cal O}_n (a\frac{n\ln \frac{1}{\rho}}{\ln n}\exp[\frac{\ln (an^2(\ln \rho)^2)}{\ln \frac{1}{\rho}} \ln \rho])\\
        &={\cal O}_n (a\frac{n\ln \frac{1}{\rho}}{\ln n}\frac{1}{an^2(\ln \rho)^2}) \\
        &={\cal O}_n(\frac{1}{n\ln(n)\ln(\frac{1}{\rho})})\\
        &=o_n(\frac{\ln n}{n\ln\frac{1}{\rho}})
    \end{align*}
    which finally gives $\gamma_K={\cal O}_n(\frac{K}{n})={\cal O}_n(\frac{\ln n}{n\ln\frac{1}{\rho}})$

    From this calculus, we can already conclude in the restart case

    $$\gamma_\mathrm{low}=\gamma={\cal O}_n(\frac{1}{n\ln(n)\ln(\frac{1}{\rho})})\qquad \gamma_\mathrm{up}={\cal O}_n(\frac{\ln n}{n\ln\frac{1}{\rho}})$$
    
    \item Let $\gamma_\mathrm{as}=\frac{\ln n}{n\ln\frac{1}{\rho}}$. Hence, we are looking for $r\in[n]$ such that asymptotically 
    \begin{itemize}
        \item $r=o(n)$ which will ensure $n-r={\cal O}_n(n)$ and also $K>>1$ and $|K\ln\rho|>>1$
        \item the remaining terms function of $r$ only are also negligible i.e $\frac{1+\alpha r}{n+1}={\cal O}_n(\gamma_\mathrm{as})$ and $\beta(r)={\cal O}_n(\gamma_\mathrm{as})$
    \end{itemize}
    
    Taking $r=\frac{\ln n}{\ln \frac{1}{\rho}}$ convenes.
\end{itemize}

Finally, we also achieve an asymptotic optimal coverage gap in the non restart case

$$\gamma_\mathrm{low}=\gamma={\cal O}_n(\frac{1}{n\ln(n)\ln(\frac{1}{\rho})})\qquad \gamma_\mathrm{up}={\cal O}_n(\frac{\ln n}{n\ln\frac{1}{\rho}})$$




 








\subsection{Proof of Proposition \ref{thm:cov_k2}}








\begin{proof}
For a given $\hat{K}_N$, we wish to compute the asymptotic coverage gap $\gamma_\mathrm{low}(\hat{K}_N)$ and $\gamma_\mathrm{up}(\hat{K}_N)$.
We start by bounding the estimation error on $\hat{K}_N$.

For $\delta\geq0$ and a universal constant $C$, let $$w=w(N,\delta)=C\sqrt{\frac{\ln \frac{|\mathcal{X}|}{\delta}\ln N}{\pi^\star(1-\rho) N}}$$

We have with probability at least $1-\delta$ 
    
    \begin{align*}
        \hat{K}_N \leq -\frac{\ln n}{\ln(\rho+w)} 
        = -\frac{\ln n}{\ln(\rho(1+\frac{w}{\rho}))} 
        &= -\frac{\ln n}{\ln(\rho)+\frac{w}{\rho}+o_N(w)} \\
        &= -\frac{\ln n}{\ln(\rho)(1+\frac{w}{\rho\ln\rho}+o_N(w))} \\
        &= -\frac{\ln n}{\ln(\rho)} (1-\frac{w}{\rho\ln\rho}+o_N(w)) \\
        &= \frac{\ln n}{\ln(\frac{1}{\rho})} +\frac{w\ln n}{\rho(\ln\rho)^2}+o_N(w) \\
        &= K +\frac{w\ln n}{\rho(\ln\rho)^2}+o_N(w) 
    \end{align*}

Similarly, $\hat{K}_N\geq K -\frac{w\ln n}{\rho(\ln\rho)^2}+o_N(w)$.

Therefore there exists a universal constant which will again denoted by $C>0$ such that

$$\mathbb{P}_\mathrm{tr}\bigg(|\hat{K}_N-K|\leq C\frac{w(N,\delta)\ln n}{\rho(\ln \rho)^2}\bigg) \geq 1-\delta$$ 

and for simplicity, we will ignore it.

For the lower bound of the coverage gap, we have with probability at least $1-\delta$

    \begin{align*}
        \gamma_\mathrm{low}(\hat{K}_N)\geq \frac{n}{K+\frac{w\ln n}{\rho(\ln\rho)^2}}\rho^{\hat{K}_N} 
        &= \frac{n}{K}(1-\frac{w\ln n}{\rho(\ln\rho)^2K}+o_N(w))\rho^K \\
        &= \gamma_\mathrm{low}-\frac{w\ln n}{\rho(\ln\rho)^2K}\gamma_\mathrm{low} +o_N(w) \\
        &= \gamma_\mathrm{low}-\frac{w}{\rho(\ln\rho)^2n\ln n} +o_N(w)
    \end{align*}

    And similarly $\hat{\gamma}_l\leq \gamma_\mathrm{low}+\frac{w}{\rho(\ln\rho)^2n\ln n} +o_N(w)$.

    Doing the same calculation for the upper bound of the coverage gap $\gamma_{up}(\hat{K}_N)=\frac{\hat{K}_N}{n}$, we finally obtain for a universal constant $C>0$
        $$\mathbb{P}_\mathrm{tr}\bigg(|\gamma_\mathrm{low}(\hat{K}_N)-\gamma_\mathrm{low}|\leq C\frac{w(N,\delta)}{\rho(\ln\rho)^2n\ln n}, |\gamma_{up}(\hat{K}_N)-\gamma_\mathrm{up}|\leq C\frac{w(N,\delta)}{n}\bigg) \geq 1-\delta$$ 
    
Let $I=\{|\gamma_\mathrm{low}(\hat{K}_N)-\gamma_\mathrm{low}|\leq C\frac{w(N,\delta)}{\rho(\ln\rho)^2n\ln n}, |\gamma_{up}(\hat{K}_N)-\gamma_\mathrm{up}|\leq C\frac{w(N,\delta)}{n}\},$ we have $\mathbb{P}_\mathrm{tr}(I)\geq 1-\delta$

By combining both probabilities, 

\begin{align*}
    \mathbb{P}(Y_{n+1}\in C_{\hat{K}_N}(X_{n+1}))&\geq \mathbb{P}(Y_{n+1}\in C_{\hat{K}_N}(X_{n+1}),I) \\
    &=\mathbb{E}_\mathrm{tr}[\mathbb{P}(Y_{n+1}\in C_{\hat{K}_N}(X_{n+1})|D_\mathrm{tr})]\\
    &\geq \mathbb{E}_\mathrm{tr}[1-\alpha-\gamma_\mathrm{low}(\hat{K}_N)]\\
    &\geq \mathbb{E}_\mathrm{tr}[1-\alpha-\gamma_\mathrm{low}-C\frac{w(N,\delta)}{\rho(\ln\rho)^2n\ln n}] \\
    &\geq 1-\alpha -\gamma_\mathrm{low}-C\frac{w(N,\delta)}{\rho(\ln\rho)^2n\ln n} - \delta
\end{align*}

and similarly for the upper bound, we have

$$\mathbb{P}(Y_{n+1}\in C_{\hat{K}_N}(X_{n+1}))\leq  1-\alpha +\gamma_\mathrm{up}+C\frac{w(N,\delta)}{n} +\delta $$

Taking $\delta=\frac{|\mathcal{X}| \ln N}{N}$, this gives $w(N,\delta)=\sqrt\frac{\ln \frac{N}{\ln N}\ln N}{\pi^\star(1-\rho)N}= {\cal O}_N(\frac{\ln N}{\sqrt{N}})$
\end{proof}

%% file: appendix-sec4.tex
\section{Experiments}

\subsection{Lazy random walk}
\label{app:lazy_walk}

For a given $w>0$, the associated transition matrix is $P:[1,w]^2\to[0,1]$ 

$$P(x,y) = 
\begin{cases}
    \frac{1}{4} & \text{if $y=x+1$ or $y =x-1$ (mod w)}, \\
    \frac{1}{2} & \text{if $y=x$},\\
    0 &\text{otherwise}.
\end{cases}$$

We consider a simple linear regression for illustration purposes and take $\mu(x)=ax$ for a certain $a\in[0,1]$. The independent noise follows $\varepsilon_t\sim N(0,1)$.

\subsection{Gaussian AR(1)}
\label{app:gaussian_ar}

It can be easily shown that the corresponding transition kernel $P$ and all its iterates are also Gaussian

$$P(a,dy)=\frac{1}{\sqrt{2\pi}\omega}\exp{(-\frac{(y-\theta a)^2}{2\omega^2})}dy$$

and

$$\forall n \quad P^n(x,dy)\sim N(\theta^n x,\omega^2\frac{1-\theta^{2n}}{1-\theta^2})$$ 

In the stationary case, we know that $\beta$ can be written as 

$$\beta(n)=\frac{1}{2}\int_\mathcal{R} \pi(dx)||P^n(x,.)-\pi||_{TV} $$

Furthermore, we also have the following Pinsker inequality on the total variation distance 

$$||P^n(x,.)-\pi||_{TV}\leq \sqrt{\frac{1}{2}\mathrm{KL}(P^n(x,.)||\pi)}$$

where $\mathrm{KL}$ is the usual KL divergence between two distributions. Or for Gaussian distributions $N(m_1,\sigma_1^2), N(m_2, \sigma_2^2)$, its expression is well known 

$$\frac{1}{2}\log\frac{\sigma_2^2}{\sigma_1^2} + \frac{\sigma_1^2 + (m_1-m_2)^2}{2\sigma_2^2}-\frac{1}{2}$$

In our case, for a given starting state $x\in\mathbb{R}$, 

\begin{align*}
    \mathrm{KL}(P^n(x,.)||\pi)&=-\frac{1}{2}\log (1-\theta^{2n}) + \frac{1-\theta^{2n}}{2}+\frac{x^2\theta^{2n}(1-\theta^2)}{\omega^2}-\frac{1}{2} \\
    &= -\frac{1}{2}\log (1-\theta^{2n})-\frac{\theta^{2n}}{2}+\frac{x^2\theta^{2n}(1-\theta^2)}{\omega^2} \\
    &\approx \frac{x^2\theta^{2n}(1-\theta^2)}{\omega^2} \quad \text{for $n>>1$}
\end{align*}

Therefore 

\begin{align*}
    \beta(n)&=\frac{1}{2}\int_\mathcal{R} \pi(dx)||P^n(x,.)-\pi||_{TV} \\
    &\leq\frac{1}{2}\int_\mathbb{R} \pi(dx)\sqrt{\frac{x^2\theta^{2n}(1-\theta^2)}{2\omega^2}} \\
    &=\frac{\theta^n}{2}\sqrt{\frac{1-\theta^2}{2\omega^2}}\int_\mathbb{R}|x|\pi(dx) \\
    &= \theta^n\sqrt{\frac{1-\theta^2}{2\omega^2}}
\end{align*}

Note also that if the initial distribution $\nu_0$ is Gaussian then $\nu_0P^n$ will also be Gaussian. Therefore given the geometric ergodicity function $Q(x)=|x|$, the condition of Theorem \ref{thm:cov_k}, $\int_\mathbb{R} Q(x)\nu_0P^N(x)dx<\infty$ will be verified for any $N$.

\newpage
\subsection{Real-world dataset}

\begin{figure}[ht]
\vskip 0.2in
\begin{center}
\centerline{\includegraphics[width=\columnwidth]{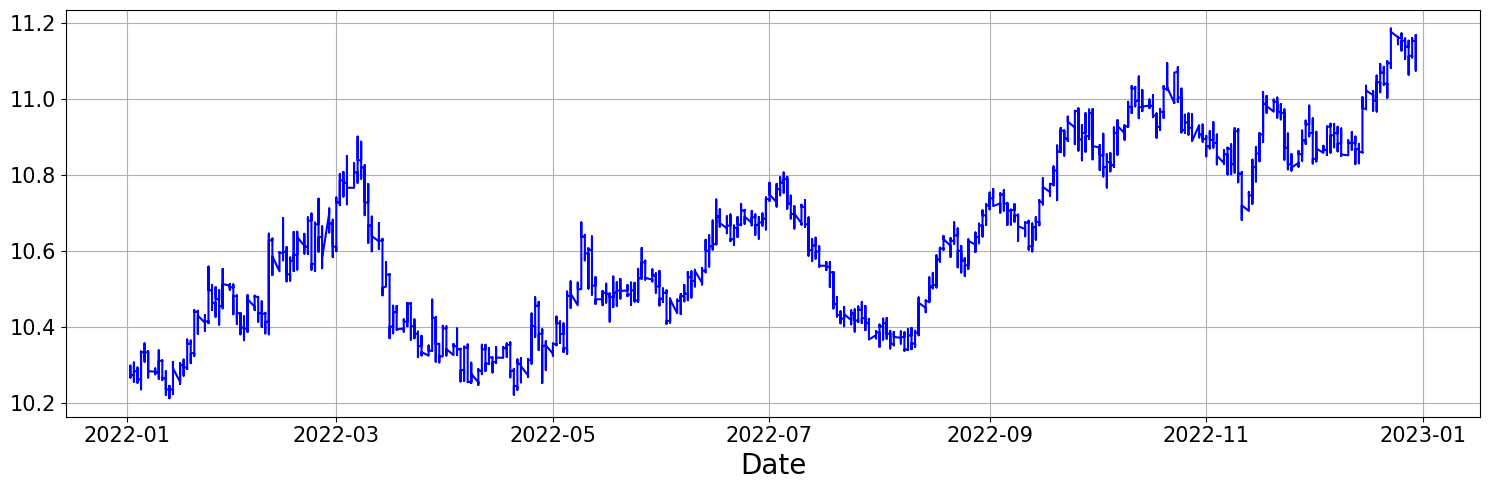}}
    \caption{Exchange rate EUR/SEK}
    \label{fig:elec}
    \end{center}
\vskip -0.2in
\end{figure}



\begin{figure}[ht]
\vskip 0.2in
\begin{center}
\centerline{\includegraphics[width=\columnwidth]{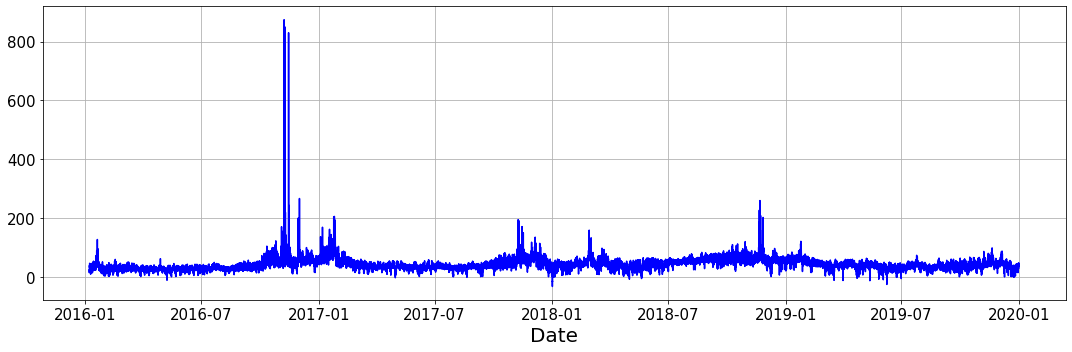}}
    \caption{French electricity price}
    \label{fig:elec}
    \end{center}
\vskip -0.2in
\end{figure}








%% file: main.bbl
\begin{thebibliography}{37}
\providecommand{\natexlab}[1]{#1}
\providecommand{\url}[1]{\texttt{#1}}
\expandafter\ifx\csname urlstyle\endcsname\relax
  \providecommand{\doi}[1]{doi: #1}\else
  \providecommand{\doi}{doi: \begingroup \urlstyle{rm}\Url}\fi

\bibitem[Angelopoulos \& Bates(2023)Angelopoulos and Bates]{angelopoulos2023}
Angelopoulos, A.~N. and Bates, S.
\newblock Conformal prediction: A gentle introduction.
\newblock \emph{Found. Trends Mach. Learn.}, 16\penalty0 (4):\penalty0 494–591, mar 2023.
\newblock ISSN 1935-8237.
\newblock \doi{10.1561/2200000101}.
\newblock URL \url{https://doi.org/10.1561/2200000101}.

\bibitem[Barber et~al.(2023)Barber, Candes, Ramdas, and Tibshirani]{barber2023conformal}
Barber, R.~F., Candes, E.~J., Ramdas, A., and Tibshirani, R.~J.
\newblock Conformal prediction beyond exchangeability.
\newblock \emph{The Annals of Statistics}, 51\penalty0 (2):\penalty0 816--845, 2023.

\bibitem[Bernstein(1927)]{Bernstein1927SurLD}
Bernstein, S.
\newblock Sur l'extension du th{\'e}or{\'e}me limite du calcul des probabilit{\'e}s aux sommes de quantit{\'e}s d{\'e}pendantes.
\newblock \emph{Mathematische Annalen}, 97:\penalty0 1--59, 1927.
\newblock URL \url{https://api.semanticscholar.org/CorpusID:122172457}.

\bibitem[Bhattacharya \& Lee(1995)Bhattacharya and Lee]{bhattacharya1995ergodicity}
Bhattacharya, R.~N. and Lee, C.
\newblock Ergodicity of nonlinear first order autoregressive models.
\newblock \emph{Journal of Theoretical Probability}, 8:\penalty0 207--219, 1995.

\bibitem[Bresler et~al.(2020)Bresler, Jain, Nagaraj, Netrapalli, and Wu]{bresler2020}
Bresler, G., Jain, P., Nagaraj, D., Netrapalli, P., and Wu, X.
\newblock Least squares regression with markovian data: fundamental limits and algorithms.
\newblock In \emph{Proceedings of the 34th International Conference on Neural Information Processing Systems}, NIPS'20, Red Hook, NY, USA, 2020. Curran Associates Inc.
\newblock ISBN 9781713829546.

\bibitem[Chatterjee(2023)]{chatterjee2023}
Chatterjee, S.
\newblock Spectral gap of nonreversible markov chains, 2023.

\bibitem[Chatzigeorgiou(2013)]{chatzigeorgiou2013bounds}
Chatzigeorgiou, I.
\newblock Bounds on the lambert function and their application to the outage analysis of user cooperation.
\newblock \emph{IEEE Communications Letters}, 17\penalty0 (8):\penalty0 1505--1508, 2013.

\bibitem[Combes \& Touati(2019)Combes and Touati]{combes2019}
Combes, R. and Touati, M.
\newblock Computationally efficient estimation of the spectral gap of a markov chain.
\newblock \emph{Proc. ACM Meas. Anal. Comput. Syst.}, 3\penalty0 (1), mar 2019.

\bibitem[Davydov(1973)]{davydov1973mixing}
Davydov, Y.~A.
\newblock Mixing conditions for markov chains.
\newblock \emph{Teoriya Veroyatnostei i ee Primeneniya}, 18\penalty0 (2):\penalty0 321--338, 1973.

\bibitem[Dixit et~al.(2023)Dixit, Lindemann, Wei, Cleaveland, Pappas, and Burdick]{dixit2023}
Dixit, A., Lindemann, L., Wei, S.~X., Cleaveland, M., Pappas, G.~J., and Burdick, J.~W.
\newblock Adaptive conformal prediction for motion planning among dynamic agents.
\newblock In Matni, N., Morari, M., and Pappas, G.~J. (eds.), \emph{Proceedings of The 5th Annual Learning for Dynamics and Control Conference}, volume 211 of \emph{Proceedings of Machine Learning Research}, pp.\  300--314. PMLR, 15--16 Jun 2023.
\newblock URL \url{https://proceedings.mlr.press/v211/dixit23a.html}.

\bibitem[Doukhan(2012)]{doukhan2012mixing}
Doukhan, P.
\newblock \emph{Mixing: Properties and Examples}.
\newblock Lecture Notes in Statistics. Springer New York, 2012.
\newblock ISBN 9781461226420.

\bibitem[Foffano et~al.(2023)Foffano, Russo, and Prouti{\`{e}}re]{foffano2023}
Foffano, D., Russo, A., and Prouti{\`{e}}re, A.
\newblock Conformal off-policy evaluation in markov decision processes.
\newblock In \emph{62nd {IEEE} Conference on Decision and Control, {CDC} 2023, Singapore, December 13-15, 2023}, pp.\  3087--3094. {IEEE}, 2023.
\newblock \doi{10.1109/CDC49753.2023.10383469}.
\newblock URL \url{https://doi.org/10.1109/CDC49753.2023.10383469}.

\bibitem[Foygel~Barber et~al.(2021)Foygel~Barber, Candes, Ramdas, and Tibshirani]{foygel2021limits}
Foygel~Barber, R., Candes, E.~J., Ramdas, A., and Tibshirani, R.~J.
\newblock The limits of distribution-free conditional predictive inference.
\newblock \emph{Information and Inference: A Journal of the IMA}, 10\penalty0 (2):\penalty0 455--482, 2021.

\bibitem[Gallegos-Herrada et~al.(2023)Gallegos-Herrada, Ledvinka, and Rosenthal]{gallegos2023equivalences}
Gallegos-Herrada, M.~A., Ledvinka, D., and Rosenthal, J.~S.
\newblock Equivalences of geometric ergodicity of markov chains.
\newblock \emph{Journal of Theoretical Probability}, pp.\  1--27, 2023.

\bibitem[Gammerman \& Vovk(2007)Gammerman and Vovk]{gammerman2007}
Gammerman, A. and Vovk, V.
\newblock Hedging predictions in machine learning.
\newblock \emph{Comput. J.}, 50\penalty0 (2):\penalty0 151–163, mar 2007.
\newblock ISSN 0010-4620.
\newblock \doi{10.1093/comjnl/bxl065}.
\newblock URL \url{https://doi.org/10.1093/comjnl/bxl065}.

\bibitem[Gibbs \& Candes(2021)Gibbs and Candes]{gibbs2021adaptive}
Gibbs, I. and Candes, E.
\newblock Adaptive conformal inference under distribution shift.
\newblock \emph{Advances in Neural Information Processing Systems}, 34:\penalty0 1660--1672, 2021.

\bibitem[Hsu et~al.(2019)Hsu, Kontorovich, Levin, Peres, Szepesvári, and Wolfer]{hsu2019}
Hsu, D., Kontorovich, A., Levin, D.~A., Peres, Y., Szepesvári, C., and Wolfer, G.
\newblock Mixing time estimation in reversible markov chains from a single sample path.
\newblock \emph{The Annals of Applied Probability}, 29\penalty0 (4):\penalty0 pp. 2439--2480, 2019.
\newblock ISSN 10505164, 21688737.

\bibitem[Hsu et~al.(2015)Hsu, Kontorovich, and Szepesv{\'a}ri]{hsu2015mixing}
Hsu, D.~J., Kontorovich, A., and Szepesv{\'a}ri, C.
\newblock Mixing time estimation in reversible markov chains from a single sample path.
\newblock \emph{Advances in neural information processing systems}, 28, 2015.

\bibitem[Kolla et~al.(2019)Kolla, Prashanth, Bhat, and Jagannathan]{kolla2019concentration}
Kolla, R.~K., Prashanth, L., Bhat, S.~P., and Jagannathan, K.
\newblock Concentration bounds for empirical conditional value-at-risk: The unbounded case.
\newblock \emph{Operations Research Letters}, 47\penalty0 (1):\penalty0 16--20, 2019.

\bibitem[Kuznetsov \& Mohri(2017)Kuznetsov and Mohri]{kuznetsov2017generalization}
Kuznetsov, V. and Mohri, M.
\newblock Generalization bounds for non-stationary mixing processes.
\newblock \emph{Machine Learning}, 106\penalty0 (1):\penalty0 93--117, 2017.

\bibitem[Lei \& Wasserman(2014)Lei and Wasserman]{lei2014distribution}
Lei, J. and Wasserman, L.
\newblock Distribution-free prediction bands for non-parametric regression.
\newblock \emph{Journal of the Royal Statistical Society: Series B: Statistical Methodology}, pp.\  71--96, 2014.

\bibitem[Lei et~al.(2018)Lei, G’Sell, Rinaldo, Tibshirani, and Wasserman]{lei2018distribution}
Lei, J., G’Sell, M., Rinaldo, A., Tibshirani, R.~J., and Wasserman, L.
\newblock Distribution-free predictive inference for regression.
\newblock \emph{Journal of the American Statistical Association}, 113\penalty0 (523):\penalty0 1094--1111, 2018.

\bibitem[Levin \& Peres(2017)Levin and Peres]{levin2017markov}
Levin, D.~A. and Peres, Y.
\newblock \emph{Markov chains and mixing times}, volume 107.
\newblock American Mathematical Soc., 2017.

\bibitem[Liebscher(2005)]{liebscher2005towards}
Liebscher, E.
\newblock Towards a unified approach for proving geometric ergodicity and mixing properties of nonlinear autoregressive processes.
\newblock \emph{Journal of Time Series Analysis}, 26\penalty0 (5):\penalty0 669--689, 2005.

\bibitem[Meyn \& Tweedie(2012)Meyn and Tweedie]{meyn2012markov}
Meyn, S.~P. and Tweedie, R.~L.
\newblock \emph{Markov chains and stochastic stability}.
\newblock Springer Science \& Business Media, 2012.

\bibitem[Nummelin \& Tuominen(1982)Nummelin and Tuominen]{nummelin1982geometric}
Nummelin, E. and Tuominen, P.
\newblock Geometric ergodicity of harris recurrent marcov chains with applications to renewal theory.
\newblock \emph{Stochastic Processes and Their Applications}, 12\penalty0 (2):\penalty0 187--202, 1982.

\bibitem[Oliveira et~al.(2022)Oliveira, Orenstein, Ramos, and Romano]{oliveira2022split}
Oliveira, R.~I., Orenstein, P., Ramos, T., and Romano, J.~V.
\newblock Split conformal prediction for dependent data.
\newblock \emph{arXiv preprint arXiv:2203.15885}, 2022.

\bibitem[Paulin(2015)]{paulin2015concentration}
Paulin, D.
\newblock Concentration inequalities for markov chains by marton couplings and spectral methods.
\newblock 2015.

\bibitem[Roberts \& Rosenthal(1997)Roberts and Rosenthal]{roberts1997geometric}
Roberts, G. and Rosenthal, J.
\newblock Geometric ergodicity and hybrid markov chains.
\newblock 1997.

\bibitem[Romano et~al.(2019)Romano, Patterson, and Candes]{romano2019}
Romano, Y., Patterson, E., and Candes, E.
\newblock Conformalized quantile regression.
\newblock In Wallach, H., Larochelle, H., Beygelzimer, A., d\textquotesingle Alch\'{e}-Buc, F., Fox, E., and Garnett, R. (eds.), \emph{Advances in Neural Information Processing Systems}, volume~32. Curran Associates, Inc., 2019.
\newblock URL \url{https://proceedings.neurips.cc/paper_files/paper/2019/file/5103c3584b063c431bd1268e9b5e76fb-Paper.pdf}.

\bibitem[Shafer \& Vovk(2008)Shafer and Vovk]{shafer2008}
Shafer, G. and Vovk, V.
\newblock A tutorial on conformal prediction.
\newblock \emph{J. Mach. Learn. Res.}, 9:\penalty0 371–421, jun 2008.
\newblock ISSN 1532-4435.

\bibitem[Tibshirani et~al.(2019)Tibshirani, Foygel~Barber, Candes, and Ramdas]{tibshirani2019}
Tibshirani, R.~J., Foygel~Barber, R., Candes, E., and Ramdas, A.
\newblock Conformal prediction under covariate shift.
\newblock In Wallach, H., Larochelle, H., Beygelzimer, A., d\textquotesingle Alch\'{e}-Buc, F., Fox, E., and Garnett, R. (eds.), \emph{Advances in Neural Information Processing Systems}, volume~32. Curran Associates, Inc., 2019.
\newblock URL \url{https://proceedings.neurips.cc/paper_files/paper/2019/file/8fb21ee7a2207526da55a679f0332de2-Paper.pdf}.

\bibitem[Vovk et~al.(2005)Vovk, Gammerman, and Shafer]{vovk2005algorithmic}
Vovk, V., Gammerman, A., and Shafer, G.
\newblock \emph{Algorithmic learning in a random world}, volume~29.
\newblock Springer, 2005.

\bibitem[Wisniewski et~al.(2020)Wisniewski, Lindsay, and Lindsay]{wisniewski2020application}
Wisniewski, W., Lindsay, D., and Lindsay, S.
\newblock Application of conformal prediction interval estimations to market makers’ net positions.
\newblock In \emph{Conformal and probabilistic prediction and applications}, pp.\  285--301. PMLR, 2020.

\bibitem[Wolfer \& Kontorovich(2019)Wolfer and Kontorovich]{wolfer2019estimating}
Wolfer, G. and Kontorovich, A.
\newblock Estimating the mixing time of ergodic markov chains.
\newblock In \emph{Conference on Learning Theory}, pp.\  3120--3159. PMLR, 2019.

\bibitem[Yu(1994)]{yu1994rates}
Yu, B.
\newblock Rates of convergence for empirical processes of stationary mixing sequences.
\newblock \emph{The Annals of Probability}, pp.\  94--116, 1994.

\bibitem[Zaffran et~al.(2022)Zaffran, F{\'e}ron, Goude, Josse, and Dieuleveut]{zaffran2022adaptive}
Zaffran, M., F{\'e}ron, O., Goude, Y., Josse, J., and Dieuleveut, A.
\newblock Adaptive conformal predictions for time series.
\newblock In \emph{International Conference on Machine Learning}, pp.\  25834--25866. PMLR, 2022.

\end{thebibliography}
